\documentclass[pdflatex,sn-mathphys-num]{sn-jnl}

\usepackage{graphicx}%
\usepackage{multirow}%
\usepackage{amsmath,amssymb,amsfonts}%
\usepackage{amsthm}%
\usepackage{mathrsfs}%
\usepackage[title]{appendix}%
\usepackage{xcolor}%
\usepackage{textcomp}%
\usepackage{manyfoot}%
\usepackage{booktabs}%
\usepackage{algorithm}
\usepackage{algpseudocode}
\usepackage{listings}%

\usepackage{makecell}

\theoremstyle{thmstyleone}%
\theoremstyle{thmstyletwo}%

\theoremstyle{thmstylethree}%

\begin{document}

\title[Article Title]{Geometric Iterative Approach for Efficient Inverse Kinematics and Planning of Continuum Robots with a Floating Base Under Environment Constraints}


\author[1]{\fnm{Ma} \sur{Congjun}}\email{scucongjun\_ma@163.com}
\author[1]{\fnm{Xiao} \sur{Quan}}\email{xiaoquan@stu.scu.edu.cn}
\author[1]{\fnm{Liu} \sur{Liangcheng}}\email{1358533912@qq.com}
\author[1]{\fnm{You} \sur{Xingxing}}\email{youxingxing@scu.edu.cn}
\author*[1,2]{\fnm{Dian} \sur{Songyi}}\email{scudiansy@scu.edu.cn}
\affil[1]{\orgdiv{College of Electrical Engineering}, \orgname{Sichuan University}, \orgaddress{\city{Chengdu}, \postcode{610065}, \country{China}}}
\affil*[2]{\orgdiv{State Key Laboratory of Intelligent Construction and Healthy Operation and Maintenance of Deep Underground Engineering}, \orgname{Sichuan University}, \orgaddress{\city{Chengdu}, \postcode{610065}, \country{China}}}


\abstract{Continuum robots with floating bases demonstrate exceptional operational capabilities in confined spaces, such as those encountered in medical surgeries and equipment maintenance. However, developing low-cost solutions for their motion and planning problems remains a significant challenge in this field. This paper investigates the application of geometric iterative strategy methods to continuum robots, and proposes the algorithm based on an improved two-layer geometric iterative strategy for motion planning. First, we thoroughly study the kinematics and effective workspace of a multi-segment tendon-driven continuum robot with a floating base. Then, generalized iterative algorithms for solving arbitrary-segment continuum robots are proposed based on a series of problems such as initial arm shape dependence exhibited by similar methods when applied to continuum robots. Further, the task scenario is extended to a follow-the-leader task considering environmental factors, and further extended algorithm are proposed. Simulation comparison results with similar methods demonstrate the effectiveness of the proposed method in eliminating the initial arm shape dependence and improving the solution efficiency and accuracy. The experimental results further demonstrate that the method based on improved two-layer geometric iteration can be used for motion planning task of a continuum robot with a floating base, under an average deviation of about 4 mm in the end position, an average orientation deviation of no more than 1°, and the reduction of average number of iterations and time cost is 127.4 iterations and 72.6 ms compared with similar methods, respectively.}




\maketitle

\section{Introduction}\label{sec1}

Continuum robots are widely used in various industries, including critical equipment maintenance \cite{bib1}, medical surgeries \cite{bib2,bib3,bib4,bib5}, and disaster rescue \cite{bib6}, due to their flexibility, safety, passive compliance, adjustable size, and ease of disengagement for breakdown situation. Inspired by the biomechanics of different animals and their organs, continuum robots employ three primary actuation methods: tendon-driven, pneumatic, and muscle-driven \cite{bib7,bib8}. The structural design of continuum robots varies significantly depending on the application. For instance, in medical surgery, where the robot interacts with the human body, concentric tube designs are commonly used \cite{bib9}. Conversely, in nuclear power plant maintenance, robots typically employ a series of universal joints \cite{bib10}. The unique combination of actuation methods and arm structures gives each continuum robot its distinct motion characteristics. While these characteristics allow robots to perform specific tasks effectively in their intended applications, they also pose significant challenges for modeling and control \cite{bib11}. Even continuum robots with similar structures and actuation methods exhibit different approaches to motion modeling, each with its own limitations and applications.

In the field of continuum robot modeling, solving inverse kinematics is one of the most researched issues \cite{bib12,bib13}. The primary objective is to rapidly and efficiently determine the actuator parameters for a multi-segment continuum robot under specified end-effector conditions and environmental factors. Existing solutions for continuum robot inverse kinematics include learning-based modeling approaches \cite{bib14,bib15,bib16,bib17,bib18}, Jacobian pseudoinverse methods \cite{bib19,bib20,bib21}, constrained optimization techniques \cite{bib22,bib23,bib24}, and metaheuristic methods based on geometric iteration \cite{bib25,bib26,bib27}. However, these methods have inherent limitations due to their specific characteristics. Learning-based methods require a high degree of consistency between training and validation data. Any discrepancy between the validation scenario and the training data can directly impact modeling accuracy. Additionally, data collection is labor-intensive and often lacks effective supervision, making it difficult to ensure the authenticity of the robot multi-dimensional data through simple post-processing. The Jacobian inverse method is widely used in rigid robotic arms, but it suffers from singularity problem during the solution process, which directly affects overall efficiency. Formulating the inverse kinematics problem as an optimization problem that integrates various constraints and objectives can address most kinematic modeling challenges for continuum robots. However, the complexity of the optimization process demands significant computational resources, especially when solving the inverse kinematics of multi-segment continuum robots under multiple constraints, which can be time-consuming. Due to the inherent curvature of the robot arm, methods based on geometric iteration have not been widely applied to continuum robots \cite{bib28}. With the advent of the virtual link model, continuum robots can now be represented by virtual links and joints in space, allowing geometric iteration methods to be gradually applied. However, most of these methods are limited to solving inverse kinematics for either a three-degree-of-freedom position or a five-degree-of-freedom pose \cite{bib29}. In 2021, the research team led by Wenfu Xu at Harbin Institute of Technology proposed the Two-Layer Geometric Iteration (TLGI) method, which was successfully applied to a four-segment tendon-driven continuum robot \cite{bib27}. This approach effectively solved the inverse kinematics problem under six-degree-of-freedom constraints and could be integrated with environmental constraints to develop corresponding motion planning solutions.

The successful application of the TLGI method has paved a new path for solving motion modeling problems in continuum robots using geometric iteration. Its key advantages include fewer iterations, shorter computation times, higher accuracy, simpler constraint integration, absence of singularities, and full coverage of end-effector degrees of freedom. However, the TLGI method still has the following problems. Since the underlying iterative strategy of the TLGI method is based on the Forward And Backward Reaching Inverse Kinematics (FABRIK) algorithm, it inherits the initial state dependency problem inherent in the FABRIK method \cite{bib30}. Additionally, the multi-mode mechanism in the TLGI method does not account for the possibility of the iterative process getting stuck in a single mode, which can lead to mode failure and local oscillation during iterations. In addition, under the environmental constraints, the TLGI method has a scarce number of candidate points for virtual joints, which is prone to the problem that the iterative solution cannot be completed within a limited number of times due to the failure of candidate points. Finally, to accommodate a broader range of continuum robot applications, motion planning problems involving floating base capabilities should also be integrated into the inverse kinematics solutions based on geometric iteration strategies.

This paper proposes an improved strategy to address the aforementioned problems in the TLGI method by adopting FABRIK for continuum robot (FABRIKc) as the core iterative approach \cite{bib29}. This results in the development of the two-layer FABRIKc (TL-FABRIKc) algorithm, based on an improved two-layer geometric iteration strategy, which is further extended to scenarios involving follow-the-leader tasks. The subsequent chapters are organized as follows: Chapter \ref{Chapter2} describes the structure, motion mechanism, and workspace analysis of continuum robots with a floating base. Chapter \ref{Chapter3} identifies the problems with existing methods, introduces the new TL-FABRIKc algorithm, and presents simulation validation and comparative analysis. Chapter \ref{Chapter4} successively describes the virtual joint updating strategy in the case of joint constraints, obstacle constraints, and follow-the-leader strategy in the case of base floating, and constructs the corresponding algorithm named TL-FABRIKc on follow-the-leader task (TL-FABRIKc-FTL) . Algorithm is validated through simulations, demonstrating the improved adaptability of continuum robots to various environments using the TL-FABRIKc-based extensions. Finally, Chapter \ref{Chapter5} introduces the experimental platform for a continuum robot with a floating base, where prototypes accuracy calibration experiments,  experiments on inverse kinematics solutions, and follow-the-leader motion under base extension are conducted. The paper concludes with a summary and findings presented in Chapter \ref{Chapter6}.

\section{Mechanism and Kinematics}
\label{Chapter2}
\subsection{Mechanism}
\label{Section2.1}
\begin{figure}[!t]
\centering
\includegraphics[width=3.4in]{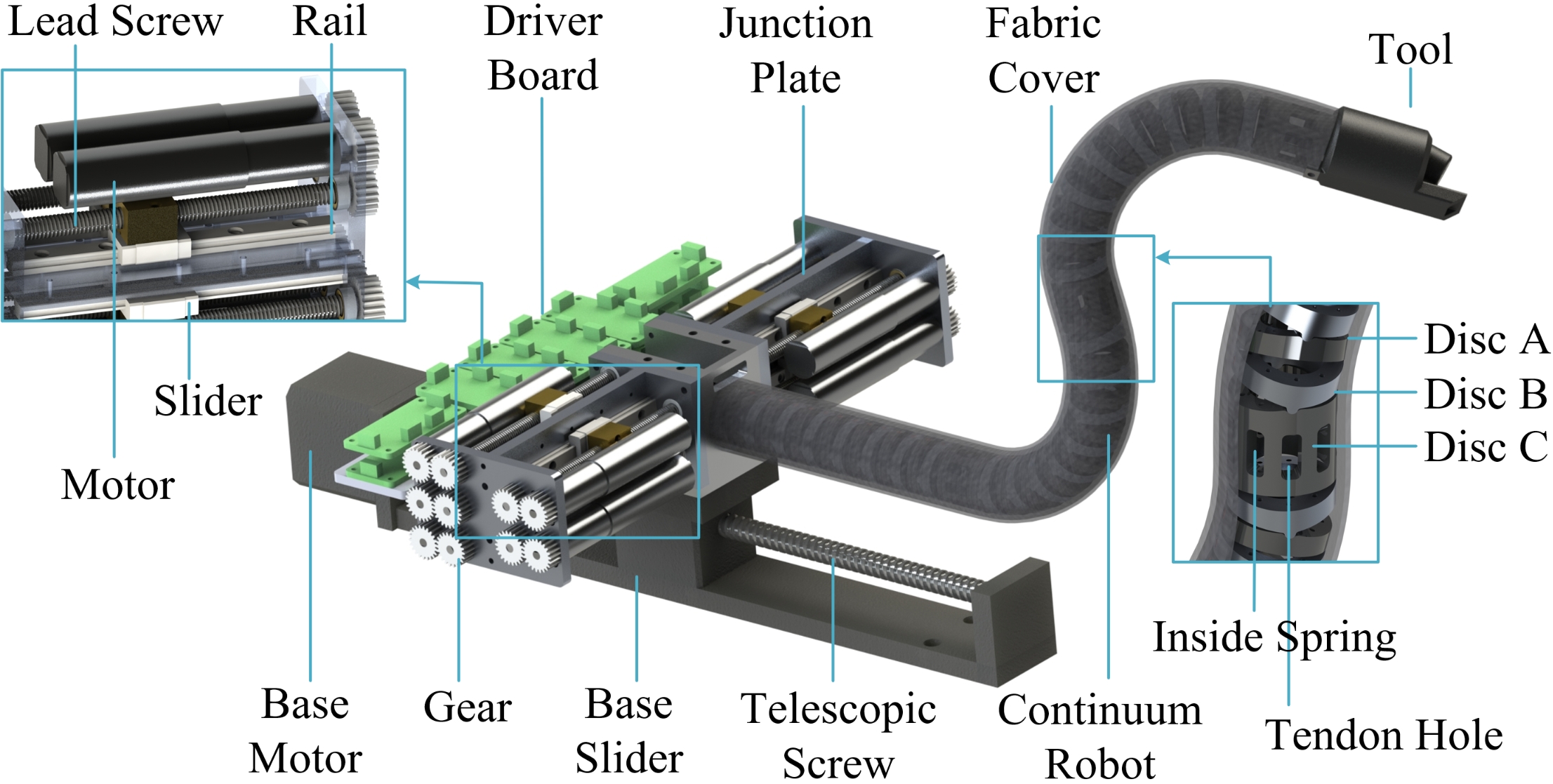}
\caption{The continuum robot with a floating base.}
\label{figure1}
\end{figure}

Fig. \ref{figure1} shows a continuum robot with a telescoping and floating base. The continuous motion of robot is achieved through tendon-driven actuation, allowing it to adapt to a wider operational range and more complex tasks with the help of the extendable floating base. The base is equipped with a motor that provides rotational motion to the telescopic lead screw, driving the slider to perform linear extension and retraction. The continuum robot illustrated in the figure consists of three segments, each controlled by three tendons that enable bending in various directions and degrees. In total, nine tendons work together to control the posture of the end-effector and the overall shape of the arm. Each tendon is anchored at one end to a micro slider, secured with fasteners. When the micro motor rotates in a specific direction, the gear set drives the micro lead screw to rotate, further moving the slider along the linear path of the screw. The tendon transmits force through the micro slider, passing through base holes and guide pulleys into the arm segment holes. The ends of each tendons are fixed to the C disk of that segment. The tendons pass through several sets of interconnected A and B disks, each containing support springs that provide structural stability to the arm. Disk A and B can rotate horizontally and vertically relative to the base around their axes of contact. For a single segment of the continuum robot, the bending direction and degree are determined by the vector sum of the rotation angles of multiple disk sets in both horizontal and vertical directions. The magnitude of the vector is proportional to the bending degree, while the vector angle relative to the reference coordinate system defines the bending angle. The end of the continuum robot is equipped with various tools suited to different operational tasks. The arm is encased in a textile fiber material to enhance safety during interaction. Table \ref{Table1} lists the design parameters of the robot.

\begin{figure}[!t]
\centering
\includegraphics[width=3.4in]{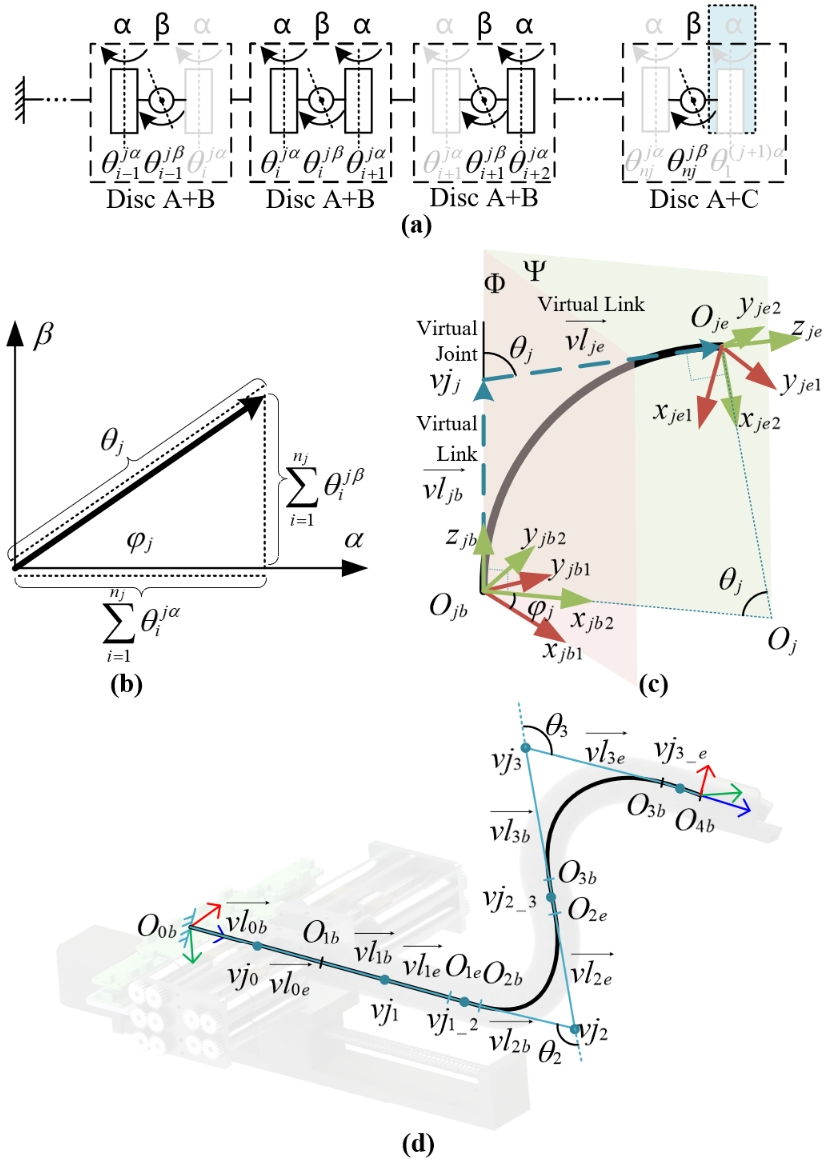}
\caption{The motion mechanism of continuum robot. (a) Joint configuration of continuum robot in ${j}^{\text{th}}$ segment. (b) Relation between axis and angle in ${j}^{\text{th}}$ segment. (c) Simplified motion mechanism of continuum robot in ${j}^{\text{th}}$ segment. (d) Virtual link model of model of continuum robot with a floating base.}
\label{figure2}
\end{figure}

\begin{table}[h]  
  \centering  
  \caption{The designed parameter of continuum robot with a floating base.} 
  \label{Table1} 
    \begin{tabular}{ll}  %
    \toprule
    \textbf{Properties} & \textbf{Value}\\
    \midrule
     Outer Diameter of Disc       & 19mm \\
     Inner Diameter of Disc        & 12.2mm \\
     Distance Between the Axes of Disk A or B & 6.37mm \\
    Distance Between the Axes of Disk C & 17.74mm \\
    Number of Disc Groups in Each Segment & 8 \\
     Disc Hole Radius        & 7.5mm \\
     Maximum Rotation Angle of Axis $\alpha$ or $\beta$ & 20° \\
    Limit of Micro Lead Screw & ±30mm \\
    Limit of Telescopic Screw & ±100mm \\
    Weight & 1.8kg \\
    Number of Segments & 3 \\
    Number of Motors & 9+1 \\
    \bottomrule
    \end{tabular}
\end{table}

\subsection{Kinematic Mechanism Analysis and Model Simplification}
\label{Section2.2}

Analysing the bending motion mechanism of the continuum robot shown in Fig. \ref{figure1} in the base-fixed case is the key to realize its motion planning. As shown in Fig. \ref{figure2}(a), which depicts the joint configuration of the ${{j}^{\text{th}}}$ segment of the robot, there are a total of ${{n}_{j}}$ groups of discs constituting, among which there are a total of ${{n}_{j}}$ discs in A, ${{n}_{j}}-1$ discs in B, and one disc in C. The joint configuration of the ${{j}^{\text{th}}}$ segment of the robot is shown in Fig.\ref{figure2}(b). The combination of discs A and B or C forms a gimbal set and provides two degrees of freedom of rotation in the horizontal and vertical directions, denoted as $\alpha $ and $\beta $ axes. The two axes transmit force through tendons to achieve passive rotation between discs. For a single-segment continuum robot, under the combined action of the tendons, the discs, and the spring, the arm bends ${{\theta }_{j}}$ degree in a specific direction ${{\varphi }_{j}}$. The bending direction and angle with respect to the angle of rotation of the individual axes between the discs satisfy the vectorial relationship depicted in Fig.\ref{figure2}(b), namely

\begin{equation}
\label{1-1}
\left\{ \begin{array}{c}
\begin{aligned}
{{\theta }_{j}} &=\sqrt{{{\left( \sum\limits_{i=1}^{{{n}_{j}}}{\theta _{i}^{j\beta }} \right)}^{2}}+{{\left( \sum\limits_{i=1}^{{{n}_{j}}}{\theta _{i}^{j\alpha }} \right)}^{2}}},0\le {{\theta }_{j}}<\pi\\
{{\varphi }_{j}} &=\tan \left( {\sum\limits_{i=1}^{{{n}_{j}}}{\theta _{i}^{j\beta }}}/{\sum\limits_{i=1}^{{{n}_{j}}}{\theta _{i}^{j\alpha }}}\; \right),0\le {{\varphi }_{j}}<2\pi\\
\end{aligned}
\end{array} \right.
\end{equation}

Where, when ${{\theta }_{j}}=0$, ${{\varphi }_{j}}=0$, the arm shape is in an initial flat state.

Experimental results in the literature \cite{bib31} show that the continuum robot of this configuration satisfies the isocurvature condition when a single segment of the arm shape is bent under the pull of the tendons, i.e., $\theta _{i}^{j\alpha }=\theta _{1}^{j\alpha },\theta _{i}^{j\beta }=\theta _{1}^{j\beta }$. Therefore, in Fig. \ref{figure2}(c), the motion state of the continuum robot at a given moment is abstracted as a fixed-length spatial arc, which lies on the plane ${\Psi} $. As shown in the figure, the bending angle ${{\theta}_{j}}$ corresponds to the central angle of the spatial arc, and the twist angle ${{\varphi }_{j}}$ is the angle between the initial plane ${\Phi} $ and the current plane ${\Psi} $. To distinguish the motion before and after deformation, Fig. \ref{figure2}(c) establishes the initial base coordinate system $\sum{{{O}_{jb1}}}$ and the current base coordinate system $\sum{{{O}_{jb2}}}$ at the center of the arc base ${{O}_{jb}}$, with green and red coordinate arrows used to differentiate them. Since the ${{z}_{jb}}$ axes of both coordinate systems overlap, only one is depicted. The initial plane ${\Phi} $ coincides with the plane ${{O}_{jb}}{{z}_{jb}}{{x}_{jb1}}$, and the current plane ${\Psi} $ coincides with the plane ${{O}_{jb}}{{z}_{jb}}{{x}_{jb2}}$. The coordinate system at the end of the ${j}^\text{th}$ arc segment $\sum{{{O}_{je}}}$ coincides with the base coordinate system of the ${j+1}^\text{th}$ arc segment $\sum{{{O}_{\left( j+1 \right)b}}}$. The transformation relationship between the tip and base of the same segment is given by the following equation:

\begin{equation}
\label{1-2}
\sum_{O_{je}}^{O_{jb}} T = \begin{bmatrix}
c\theta + s^2 \varphi k & -s\varphi c\varphi k & \frac{l}{\theta} c\varphi k & c\varphi s\theta \\
-c\varphi s\varphi k & c\varphi + c^2 \varphi k & \frac{l}{\theta} s\varphi k & s\varphi s\theta \\
-c\varphi s\theta & -s\varphi s\theta & \frac{l}{\theta} s\theta & c\theta \\
0 & 0 & 0 & 1
\end{bmatrix}
\end{equation}

Here, $k=1-\operatorname{c}\theta $, where $cos$ is simplified as $c$, and $sin$ as $s$. The subscripts of $\theta ,\varphi ,l$ are omitted for clarity. ${l}$ represents the length of the ${{j}^{\text{th}}}$ segment of the arm.

The forward kinematics model of the entire robot can be derived using the transformation matrices between adjacent segments. However, constructing an inverse kinematics model is challenging because solving the complex trigonometric inverse functions for an analytical solution is difficult. Therefore, applying transformation matrices to the arc model for inverse kinematics is problematic. To address this, in Fig. \ref{figure2}(c), the arc model is further simplified into a basic linkage model composed of virtual links and virtual joints. In this model, the arc is replaced by virtual links $\overrightarrow{v{{l}_{jb}}},\overrightarrow{v{{l}_{je}}}$ and a virtual joint $v{{j}_{j}}$. According to the geometric relationships of the original arc model, the lengths of the two virtual links are equal, and they satisfy the following relationship:

\begin{equation}
\label{1-3}
\left| \overrightarrow{v{{l}_{jb}}} \right|=\left| \overrightarrow{v{{l}_{je}}} \right|=\frac{{{l}_{j}}}{{{\theta }_{j}}}\tan \frac{{{\theta }_{j}}}{2}
\end{equation}

The directions of the virtual links align with $\overrightarrow{z{{j}_{b}}},\overrightarrow{z{{j}_{e}}}$ respectively. Therefore, the two virtual links can also be expressed as vectors:

\begin{equation}
\label{1-4}
\left\{ \begin{array}{c}
\overrightarrow{v{{l}_{jb}}}=\overrightarrow{{{z}_{jb}}}\left| \overrightarrow{v{{l}_{jb}}} \right| \\ 
\overrightarrow{v{{l}_{je}}}=\overrightarrow{{{z}_{je}}}\left| \overrightarrow{v{{l}_{je}}} \right|\\ 
\end{array} \right.
\end{equation}

Therefore, it is not difficult to obtain the coordinates of the virtual joint $v{{j}_{j}}$ point as,

\begin{equation}
\label{1-5}
v{{j}_{j}}={{O}_{jb}}+\overrightarrow{v{{l}_{jb}}}={{O}_{je}}-\overrightarrow{v{{l}_{je}}}
\end{equation}

\begin{figure*}[!t]
\centering
\includegraphics[width=5in]{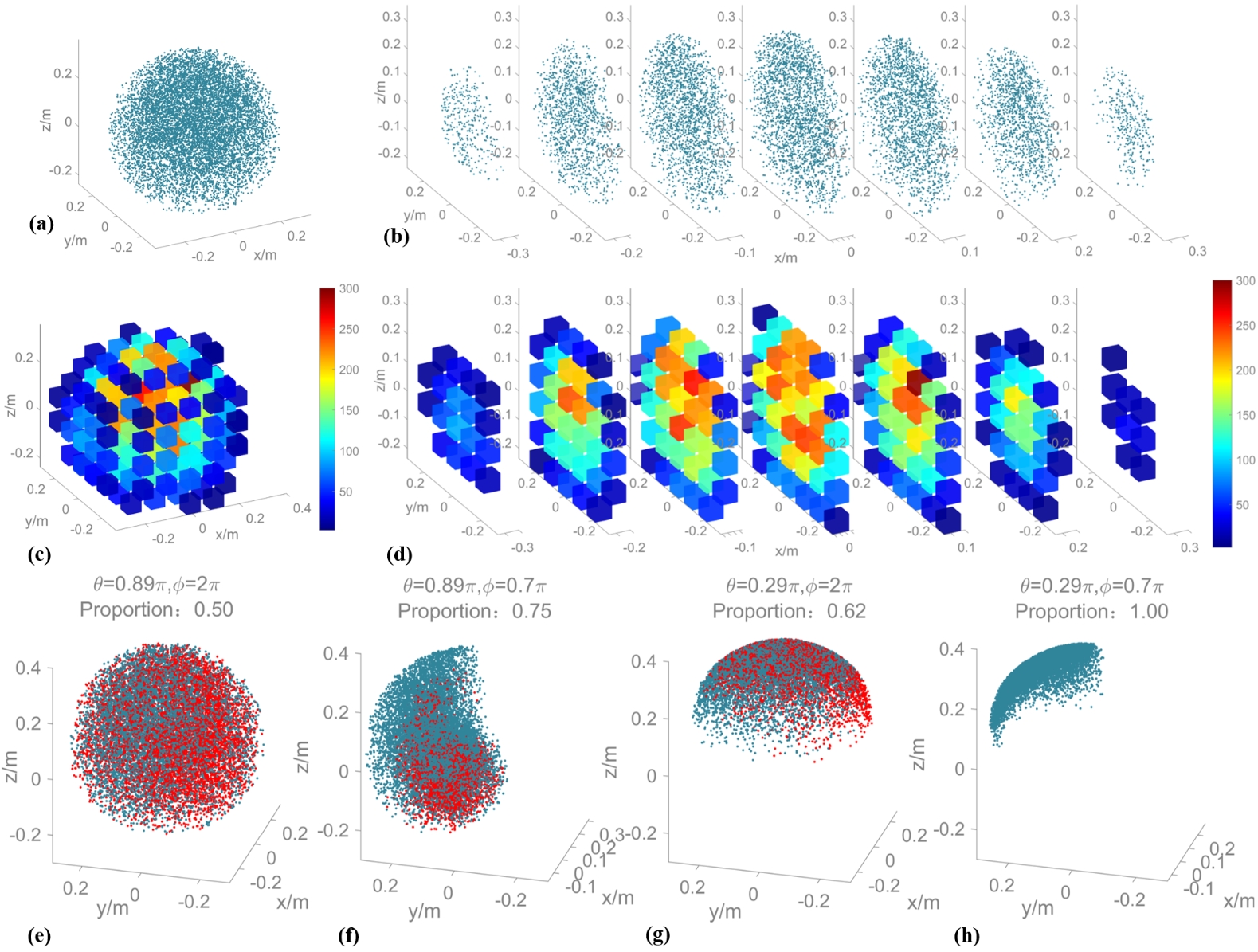}
\caption{The workspace of continuum robot. (a) Global workspace. (b) Global workspace (Layered along the $x$-axis). (c) Valid workspace. (d) Valid workspace (Layered along the $x$-axis). (e) Workspace under limitations of tendons when $\theta=0.89\pi, \phi=2\pi$. (f) Workspace under limitations of tendons when $\theta=0.89\pi, \phi=0.7\pi$.  (g)Workspace under limitations of tendons when $\theta=0.29\pi, \phi=2\pi$. (h) Workspace under limitations of tendons when $\theta=0.29\pi, \phi=0.7\pi$. }
\label{figure3}
\end{figure*}

Furthermore, based on the principles of the virtual link model, a virtual link model of the continuum robot with a floating base, as shown in Fig. \ref{figure2}(d), was developed. This model serves as the foundation for studying motion planning problems for continuum robots using geometric iteration methods. To enhance modelling accuracy during the transition from the arc model to the virtual link model, the disk C is treated as a straight segment of fixed length in the figure. The virtual joint corresponding to the disk C at the end of  the
${j}^{th}$  segment is denoted as $v{{j}_{j\_j+1}}$. The floating base, which has telescopic freedom, is considered a straight segment with variable length. The virtual links and virtual joints are represented as $\overrightarrow{v{{l}_{0b}}},\overrightarrow{v{{l}_{0e}}},v{{j}_{0}}$ respectively.

\subsection{Workspace analysis}
\label{Section2.3}

Analyzing the end-effector workspace of continuum robots is crucial for guiding their practical application in real-world scenarios. Fig. \ref{figure3}(a) illustrates the overall workspace of its end-effector, simulated based on the design parameters listed in Table \ref{Table1}. The maximum bending angle of each of the three flexible arms was set to the maximum achievable angle for a single segment, 20° for 8 groups, equivalent to $0.89\pi $. The simulation was run 5000 times, using the forward kinematic model derived from Equation (2) to compute the overall workspace. Due to the dense overlap in the spatial scatter plot and the fact that the floating base only affects the workspace along the $z$-axis, variations in the floating base were ignored during the calculation. Additionally, to facilitate the analysis of the distribution of workspace points, Fig. \ref{figure3}(b) provides a layered visualization of the overall workspace along the x-axis with intervals of 0.1m, based on Fig. \ref{figure3}(a). It can be observed that the workspace near the first and second arm segments has limited accessibility, while the region surrounding the third segment is densely populated with reachable points.

Since scatter plots only show the positions that the robot end-effector can reach, without indicating whether those points are suitable for performing tasks, determining the effective workspace that includes end-effector pose information is key to enhancing the practical utility of continuum robots. In Fig. \ref{figure2}(c), the overall workspace from Fig. \ref{figure2}(a) is divided into groups using cubic cells with a 0.1m edge length. Each cell is color-coded based on the number of end-effector points and the directionality within the cell. Higher values indicate a greater number of points and more significant directional variation, suggesting the ability to perform diverse tasks within a confined space. Equation (6) provides the formula for calculating the regional workspace effectiveness metric $S$, which quantifies the number of points $N$ and the directional variation, corresponding to the color map,

\begin{equation}
\label{1-6}
S=N\cdot \left( 1+{D}/{{{D}_{\max }}}\; \right)
\end{equation}

Among them, for the robot arm, the combination of end position ${{O}_{4e}}$ and orientation ${{z}_{4e}}$ has been able to adapt to most operational scenarios, such as fixed-point viewing, linear sweeps and area work, etc. Therefore, $D$ in Equation (6) can be calculated by the angle ${{\delta }_{{{v}_{1}}{{v}_{2}}}}$ between the orientation vectors of the end working points.

\begin{equation}
\label{1-7}
D=\frac{1}{N\left( N-1 \right)}\sum\nolimits_{{{v}_{1}}=1}^{N}{\sum\nolimits_{{{v}_{2}}={{v}_{1}}+1}^{N}{{{\delta }_{{{v}_{1}}{{v}_{2}}}}}}
\end{equation}

${{D}_{\max }}$ represents the maximum degree of dispersion, defined as the level of dispersion when the orientation vectors are completely random, set to $0.5\pi $. Additionally, the calculation of the angles between vectors is only performed when the number of workspace points in a region satisfies $N\ge 2$.

After computing the metric $S$ for each small cubic cell, a color map is used to visualize the effective workspace of the continuum robot, with equidistant layers along the $x$-axis, as shown in Fig. \ref{figure2}(c) and (d). The optimal working region of the end-effector is located within a cube with an edge length of 0.4m inside the workspace. The further away from this region, the lower the value of the evaluation metric $S$.

Table \ref{Table1} specifies that the effective stroke of the micro lead screw is ±30mm. This indicates that reaching certain workspace points in Fig. \ref{figure3}(a) may require a stroke exceeding this limit. Equation (8) presents the relationship between the length change of the nine drive lines and the bending and rotation angles of each segment. The calculated results serve as the criteria for determining whether a workspace point is reachable, and a workspace map considering the lead screw stroke is plotted in Fig. \ref{figure2}(e) and (f). In the figures, red scatter points represent points that do not meet the stroke constraints. In Equation(8), $m$ denotes the ${m}^{\text{th}}$ drive line of the  ${j}^{\text{th}}$  segment, with $m=1,2,3$. The index $jj$ reflects the coupling relationship between multiple segments, meaning that when calculating the chord length change of the  ${m}^{\text{th}}$ tendon, the effects of bending and rotation on each segment must be considered simultaneously. To understand the impact of the maximum values of bending and rotation angles on the proportion of reachable points, the maximum bending angle for the three segments is set to $0.89\pi $ or $0.29\pi $, and the maximum rotation angle is set to $2\pi $ or $0.7\pi $, with various combinations tested. The results indicate that appropriately reducing the range of rotation angles helps decrease the proportion of unreachable points under the limited stroke, although this limits the range of motion in space. However, the high density of reachable points within a unit space significantly enhances the value of the regional effectiveness metric $S$, and combined with the role of the floating base, it allows the robot to adapt to more complex task scenarios within a larger spatial range.

\section{The Proposal and Simulation Analysis of TL-FABRIKc Algorithm}
\label{Chapter3}
\subsection{Problem Formulation}
\label{Section3.1}

The FABRIK method, which is based on geometric distance, offers high accuracy, avoids singularities, and is computationally efficient for solving the inverse kinematics of serial-link robots through rapid iterations. By converting the mechanism model into a link model, this method can also be applied to solve the inverse kinematics of continuum robots with continuous bending properties. The TLGI algorithm refines the FABRIK method to address the inverse kinematics of continuum robots. Its primary advantage is the implementation of three distinct operational modes, which overcome the limitations of the traditional FABRIK algorithm in achieving six degrees of freedom for the end-effector. Consequently, the TLGI algorithm is currently the first effective method based on geometric iterative strategies that solves the inverse kinematics of multi-segment continuum robots given a specified end-effector position and orientation. However, when applying the TLGI method to solve the inverse kinematics of continuum robots, several problems remain, including dependency on the initial state, potential failure of multiple operational modes, inability to avoid obstacles, and limitation to fixed-base scenarios. This paper addresses these challenges by analyzing the existing method and proposing the new TL-FABRIKc algorithm, which enhances the capability for the inverse kinematics of continuum robots.

 \begin{figure*}[!t] 
 	\centering
 	\begin{equation}
\label{1-8}
\Delta l_{j}^{m}=\sum\nolimits_{jj=1}^{j}{r{{\theta }_{jj}}\cos \left( {{\varphi }_{jj}}+{40\pi \left( j-1 \right)}/{180}\;+2\pi {\left( m-1 \right)}/{3}\; \right)}
\end{equation}
 \end{figure*}
 
\begin{figure}[h]
\centering
\includegraphics[width=3.5in]{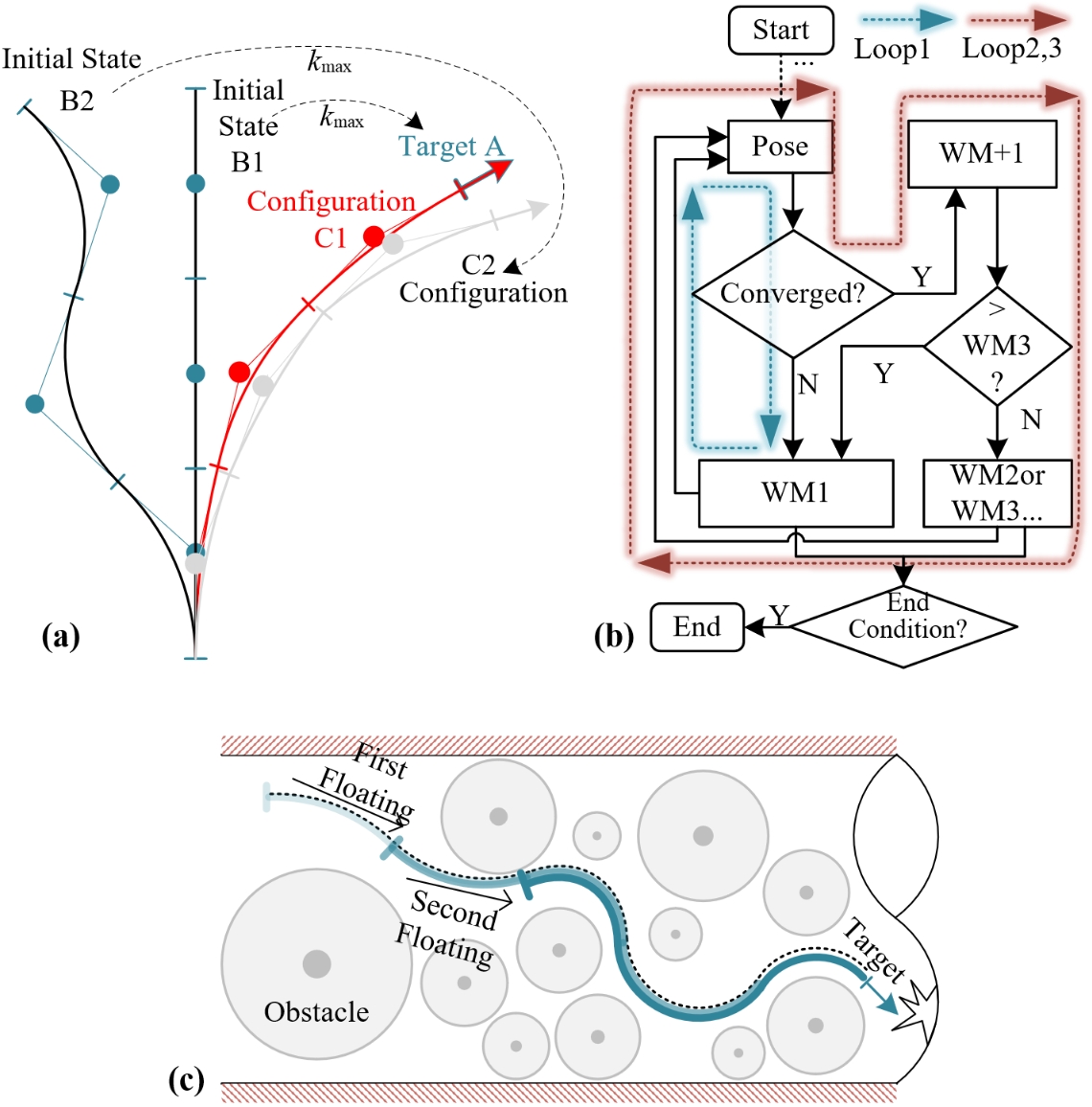}
\caption{Problems description. (a) Illustration of \textit{Problem} 1. (b) Illustration of \textit{Problem} 2. (c) Illustration of \textit{Problem} 3.}
\label{figure4}
\end{figure}

\textit{Problem 1: Dependence on Initial State.}
The TLGI algorithm fundamentally relies on the results of each forward iteration to update the positions of the virtual joints and the lengths of the virtual links in each robot segment. Similarly, each backward iteration depends on the forward iteration to update the parameters. Consequently, the initial state information propagates through the entire iterative process. This dependence on the initial configuration can lead to scenarios where, given a target state A, the solution is valid when starting from an initial state B1, but invalid when starting from a different initial state B2. As illustrated in Fig. \ref{figure4}(a), starting from B1 and iterating toward target A results in a configuration C1 that meets the target requirements. However, starting from B2 and iterating toward A yields a configuration C2, which does not satisfy the target. Further iterations continue to oscillate between similar configurations, resulting in a failure to reach the solution within the allowed number of iterations.

\textit{Problem 2: Failure of Multi-workmode.}
The TLGI algorithm employs three operational modes to address the inverse kinematics of a continuum robot under six degrees of freedom constraints. Workmode 1, 2, and 3 involve rotating the base, the end-effector, or both around the $z$-axis by a specified angle before applying the modified FABRIK algorithm. The mode is then adjusted based on the iteration results until the target pose constraints are met. However, when the local iteration oscillation described in \textit{Problem} 1 occurs, the results of each iteration fail to meet the conditions required to transition to the next operational mode. This leads to all iterations being consumed in the initial modes, as depicted in Fig. \ref{figure4}(b), where the solution process is trapped in loops 1, 2, or 3, preventing the completion of the task within the allowed number of iterations.

\textit{Problem 3: Incompatibility with a Floating Base.}
Considering the environmental factors in the TLGI algorithm, the updating strategy of the virtual joints relies on the arm shape of the previous iteration, and only a very small number of candidate points for virtual joints are obtained, which directly affects the iteration results under the environmental constraints. Therefore, it is necessary to jump out of the constraint limitation of the initial arm shape on the virtual joints at an appropriate time during the iteration process of the complex obstacle scene. In addition, increasing the number of segments in a continuum robot significantly enhances its adaptability to complex environments, but also introduces challenges in manufacturing and control. To address these challenges, continuum robots with floating bases shift the degrees of freedom that are difficult to achieve in the arm mechanism to the platform base, thereby increasing the overall maneuverability and environmental adaptability. As illustrated in Fig. \ref{figure4}(c), the robot achieves targeted inspection of a tubular space by combining two base movements with adjustments in arm configuration. However, existing geometric iteration-based methods are designed for fixed-base scenarios. Research into adapting these methods for floating-base continuum robots is crucial for tasks such as follow-the-leader motion planning.

\subsection{TL-FABRIKc Algorithm Based on Improved Two-layer Geometric Iteration}
\label{Section3.2}

The TLGI algorithm consists of inner-loop iteration and outer-loop iteration, in which the inner-loop iteration is an iteration algorithm improved on the basis of FABRIK, aiming at completing the inverse kinematics solution under the constraints of five degrees of freedom at the end of the continuum robot. The outer-loop iteration is for the rotational deviation of the $z$-axis at the end of the robot, which provides three different working modes and combines with the inner-loop iteration to complete the inverse kinematics solution problem of the continuum robot. The TL-FABRIKc algorithm is based on the TLGI algorithm, adopts FABRIKc as the inner-loop iteration method, and utilizes the improved two-layer iteration strategy to solve the inverse kinematics problem under the constraints of six degrees of freedom at the end of the continuum robot. Where TLGI uses an inner-loop iteration strategy consisting of three segments, namely forward reaching (FR) iteration, backward reaching (BR) iteration, and parameter updating (PU), FABRIKc combines the parameter updating process with the iterative process, and thus has only two segments, FR and BR. Algorithm \ref{Algorithm1} lists the pseudo-code of the TL-FABRIKc algorithm.

\begin{figure*}[!t] 
\centering
\small
\begin{equation}
\label{1-9}
 \begin{array}{c}
Ca =\operatorname{CJ}\left( e({{k}_{iter1}}),e({{k}_{iter1}}-1),e({{k}_{iter1}}-2) \right) \\ 
   = \left( abs\left( e\left( {{k}_{iter1}}-1 \right)-e\left( {{k}_{iter1}} \right) \right)\le \varepsilon  \right)\And\And\left( abs\left( e({{k}_{iter1}}-2)-e({{k}_{iter1}}-1) \right)\le \varepsilon  \right)\parallel  \\ 
     \left( abs\left( e\left( {{k}_{iter1}}-1 \right)-abs\left( e\left( {{k}_{iter1}} \right) \right) \right)\le 0 \right)\And\And\left( abs\left( e\left( {{k}_{iter1}}-2 \right)-abs\left( e\left( {{k}_{iter1}}-1 \right) \right) \right)\le 0 \right) \\

  \end{array}
\end{equation}
 \end{figure*}

For a continuum robot with an initial arm shape $\theta _{i}^{i},\varphi _{i}^{i}$ and a segment length ${{L}_{i}}$, given the target end pose ${{T}_{e}}$, find the target arm shape $\theta _{i}^{t},\varphi _{i}^{t}$ which satisfies the condition. First, the initial arm shape is converted from the circular arc model to the link model using the ARC2LINK function and entered into the loop as the parameters of the link model, and the deviation of the current position from the target position is calculated at the end of each loop to decide whether the loop continues or not. In each cycle, only any one of the workmode 1-4 can be entered, of which workmode 1-3 are proposed in the TLGI algorithm, and workmode 4 is proposed in this paper for the aforementioned \textit{Problem} 1. In workmode 4, the iterative algorithm will re-generate a random arm shape that satisfies the joint constraints as the new initial arm shape and re-invoke the TL-FABRIKc algorithm. The new randomized arm shape will be adopted in time in the early stage of falling into local iterative oscillations, thus avoiding the problem of dependence on the original initial arm shape that occurs in \textit{Problem} 1 and improving the solution success rate of the algorithm.

\begin{algorithm}[H]
\caption{Improved two-layer geometric iteration strategy based on FABRIKc. (TL-FABRIKc)}\label{Algorithm1}
\begin{algorithmic}[1]
\Require{Initial Shape $\theta_{i}^{i}, \varphi_{i}^{i}$; Ideal Tip Pose ${{T}_{e}}$; Segment Length ${{L}_{i}}$; Max Iteration Number ${{k}_{\max1}}$; Max Iteration Number of Workmode 4 $k_{\max1}^{w4}$; Max Randomization Number ${{k}_{\max2}}$; Proportion of Inner-loop Iteration Assignments ${{p}_{wm}}$; Allowable Tip Position Error ${{e}_{\min}}$; Workmode ${{w}_{c}}$}
\Ensure{Shape Configuration $\theta _{i}^{t},\varphi _{i}^{t}$}
\State \textbf{Initialization Parameter}: ${{k}_{iter1}}=0$, ${{k}_{wm}}=[0, 0, 0, 0]$, $w=1$, $result=1$
\State ${{P}_{link}} = \Call{ARC2LINK}{\theta _{i}^{i},\varphi _{i}^{i},{{L}_{i}}}$
\State $e(0) = \Call{ERR}{{{P}_{link}},{{T}_{e}}}$
\While{$e({{k}_{iter1}}) \ge {{e}_{\min}}$}
    \If{${{k}_{iter1}} < {{k}_{\max 1}}$}
\If{$w == 1$ \textbf{or} $w == 2$ \textbf{or} $w == 3$}
    \State ${{k}_{wm}}(w) = {{k}_{wm}}(w) + 1$
    \State ${{P}_{link}} = \Call{WM}{w, {{P}_{link}}}$
\ElsIf{$w == 4$}
    \State ${{k}_{wm}}(w) = {{k}_{wm}}(w) + 1$
    \State $(\theta _{i}^{r},\varphi _{i}^{r}) = \Call{RAND}{\theta _{i}^{i},\varphi _{i}^{i}}$
    \State $(\theta _{i}^{*},\varphi _{i}^{*}) = \Call{TL-FABRIKc}{\theta _{i}^{r}, \varphi _{i}^{r}, {{T}_{e}}, {{L}_{i}}, k_{\max 1}^{w4}, {{k}_{\max 2}}, {{p}_{wm}}, {{e}_{\min}}, 4}$
    \State ${{P}_{link}} = \Call{ARC2LINK}{\theta _{i}^{*},\varphi _{i}^{*},{{L}_{i}}}$
\EndIf
        \State $e({{k}_{iter1}}) = \Call{ERR}{{{P}_{link}},{{T}_{e}}}$
        \State ${{C}_{a}} = \Call{CJ}{e({{k}_{iter1}}), e({{k}_{iter1}}-1), e({{k}_{iter1}}-2)}$
        \State ${{C}_{b}} = {{k}_{wm}}(w) \ge {{p}_{wm}}(w) \cdot {{k}_{\max 1}}$
        \State $w = w + \Call{TONUM}{{{C}_{a}}}$
        \State $w = w + \Call{TONUM}{{{C}_{a}} == 0 \land {{C}_{b}}}$
        \If{$w == {{w}_{c}}$}
            \State $w = 1$
            \State ${{k}_{wm}} = [0, 0, 0, 0]$
        \EndIf
    \Else
        \State $result = 0$
        \State \textbf{break}
    \EndIf
    \State ${{k}_{iter1}} = {{k}_{iter1}} + 1$
    \State $e({{k}_{iter1}}) = \Call{ERR}{{{P}_{link}},{{T}_{e}}}$
\EndWhile
\State $(\theta _{i}^{t},\varphi _{i}^{t}) = result \cdot \Call{LINK2ARC}{{{P}_{link}}}$
\State \textbf{return} $\theta_{i}^{t},\varphi_{i}^{t}$
\end{algorithmic}
\end{algorithm}

Whether to enter or exit a certain working mode is jointly determined by conditions \(Ca\) and \(Cb\), where condition \(Ca\) reflects the iteration convergence, defined as Equation (9), where $\varepsilon$ denotes the threshold for determining that the difference of the errors in the results of the two adjacent iterations reaches the convergence condition. Condition \(Cb\) reflects whether the number of iterations in the current operating mode reaches the highest number assigned ${{p}_{wm}}(w)\cdot {{k}_{\max 1}}$. In the TLGI algorithm, due to the absence of condition \(Cb\), when the local iterative oscillation phenomenon in \textit{Problem} 1 occurs, the result of condition \(Ca\) is always false, and the WM\(w\) remains unchanged, resulting in the remaining number of iterations ${{k}_{\max 1}}-{{k}_{iter1}}$ that will be fully dissipated in the current workmode. Therefore, condition \(Cb\) is the key condition to break the phenomenon of localized iterative oscillations, and also to avoid the failure of the multiple workmode mechanism, which can effectively solve \textit{Problem} 2 and dramatically improve the success rate of the solution.

\begin{figure*}[!t]
\centering
\includegraphics[width=5in]{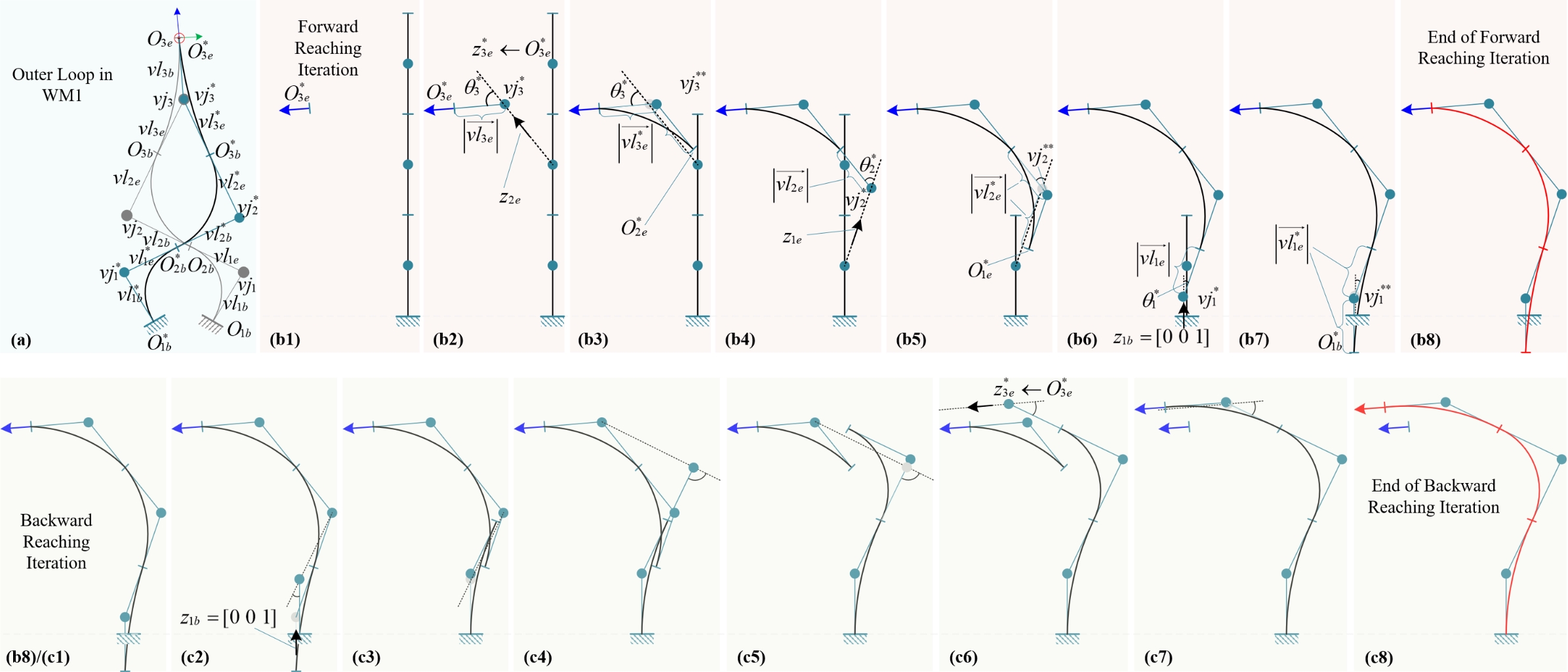}
\caption{The mechanism of TL-FABRIKc under WM1. (a) Outer loop in WM1. (b) Forward reaching iteration in WM1. (c) Backward reaching iteration in WM1.}
\label{figure5}
\end{figure*}

Workmode 1-3 involve three similar iterative processes that incorporate the FABRIKc algorithm. Fig. \ref{figure5} uses Work Mode 1 (WM1) as an example to illustrate the fundamental principles of these iterations. Among them, Fig. \ref{figure5}(a) represents the outer loop iteration process, and Figs. \ref{figure5}(b) and (c) represent the forward and backward iteration processes of the inner loop iteration, respectively. The initial arm configuration is defined with $\Sigma {{O}_{1b}}$ as the base and $\Sigma {{O}_{3e}}$ as the end-effector pose. The virtual joints, virtual links, and segment nodes can be transformed from the ARC2LINK using the parameters $\theta _{i}^{i},\varphi _{i}^{i},{{L}_{i}}$. Upon entering WM1, a forward iteration of FABRIKc is performed first, as shown in Fig. \ref{figure5}(b1)-(b8), to align the end-effector orientation with the target orientation, ensuring ${{z}_{3e}}=z_{3e}^{*}$. Subsequently, the initial arm configuration is rotated about this orientation axis to correct the deviation, resulting in a new arm configuration with $\Sigma O_{1b}^{*}$ as the base and $\Sigma O_{3e}^{*}$ as the end-effector pose in Fig. \ref{figure5}(a). Since the new arm end matches the target pose while the base is offset, a backward iteration of FABRIKc, shown in Fig. \ref{figure5}(c1)-(c8), is required again to realign the base with its original position. Finally, a complete FABRIKc iteration is executed to complete one full cycle of WM1, as depicted in Fig. \ref{figure5}(b1)-(c8). For WM1, its whole iteration process is shown in Fig. \ref{figure5}(b), Fig. \ref{figure5}(c), Fig. \ref{figure5}(a), Fig. \ref{figure5}(b), and Fig. \ref{figure5}(c).

The updated parameters in Fig. \ref{figure5}(a) are calculated as follows:

\begin{equation}
\label{1-9}
{{p}^{*}}=\mathbf{R}\left( p-{{O}_{3e}} \right)+{{O}_{3e}}
\end{equation}

Where $p$ denotes the coordinates of the points on the original arm shape and ${{p}^{*}}$ denotes the coordinates of the corresponding points on the new arm shape. The matrix $\mathbf{R}$ denotes the rotation matrix corresponding to the process of rotating the angle $\Delta \theta$ around ${{z}_{3e}}$:

\begin{equation}
\label{1-10}
\begin{array}{c}
\begin{aligned}
 \mathbf{R} &=\text{Rot}\left( {{z}_{3e}},\Delta \theta  \right) \\ 
  &=c\Delta \theta \cdot \mathbf{I}+\left( 1-c\Delta \theta  \right){{z}_{3e}}z_{3e}^{\text{T}} \\ 
 &+\operatorname{s}\Delta \theta \left[ \begin{matrix}
   &&0  &&-{{z}_{3e}}\left( 3 \right)  &&{{z}_{3e}}\left( 2 \right)  \\
   &&{{z}_{3e}}\left( 3 \right) &&0  &&-{{z}_{3e}}\left( 3 \right)  \\
   &&-{{z}_{3e}}\left( 2 \right)  &&{{z}_{3e}}\left( 1 \right)  &&0  \\
\end{matrix} \right]  
\end{aligned}
\end{array}.  
\end{equation}

In Fig.\ref{figure5}(b1), the unit vector where the virtual link $v{{l}_{3e}}$ located is determined from the given end position $\Sigma O_{3e}^{*}$ as ${{z}_{3e}}=\Sigma O_{3e}^{*}\left( 1:3,3 \right)$. In Fig.\ref{figure5}(b2), the length of the original virtual link $\left| \overrightarrow{v{{l}_{3e}}} \right|$ is extended along the reverse direction of that unit vector to the new virtual joint $vj_{3}^{*}=O_{3e}^{*}-{{z}_{3e}}\cdot \left| \overrightarrow{v{{l}_{3e}}} \right|$. Connecting the new virtual joint $vj_{3}^{*}$ to the original virtual joint $v{{j}_{2}}$, a new pinch angle $\theta _{3}^{*}$ formed by the vector ${{z}_{2e}}={\overrightarrow{v{{j}_{2}}vj_{3}^{*}}}/{\left| \overrightarrow{v{{j}_{2}}vj_{3}^{*}} \right|}\;$ and ${{z}_{3e}}$, which is the 3rd bending angle under the current iteration.

\begin{equation}
\label{1-11}
\theta _{3}^{*}=\text{acos}\left( {{z}_{3e}}\cdot {{z}_{2e}} \right)
\end{equation}

After obtaining the new bending angle, the length of the virtual linkage is updated to $\left| \overrightarrow{vl_{3e}^{*}} \right|$ in Fig.\ref{figure5}(b3), and the virtual joints are updated again, i.e., $vj_{3}^{**}$.

\begin{equation}
\label{1-12}
\left| \overrightarrow{vl_{3e}^{*}} \right|=\frac{{{L}_{3}}}{\theta _{3}^{*}}\tan \frac{\theta _{3}^{*}}{2}
\end{equation}

So far, all the parameters of the end segments in the forward iteration complete update, followed by the parameter updates of the second segment in Fig.\ref{figure5}(b4)-(b5) and the first segment in Fig.\ref{figure5}(b6)-(b7). Noteworthy, ${{z}_{1b}}=[0,0,1]$ is the default value for the fixed-base case when performing the parameter update in the first segment. Finally, the forward iteration is completed to form the new arm shape in Fig.\ref{figure5}(b8).

In the backward iteration phase, the first step involves updating the parameters of the segment closest to the base. At this stage, the original base pose $\Sigma {{O}_{1b}}$ is known and considered equivalent to the end-effector target pose $\Sigma O_{3e}^{*}$ from the forward iteration. Therefore, the parameter update in the backward iteration follows the same principles as in the forward iteration. It is also important to note that during the parameter update of the final segment, the end direction ${{z}_{3e}}$ still satisfies ${{z}_{3e}}=\Sigma O_{3e}^{*}\left( 1:3,3 \right)$. Upon completing the backward iteration, the final arm configuration is formed as shown in Fig.\ref{figure5}(c8). Since there may still be a difference in the end-effector pose, the process enters another iteration cycle. This cycle continues until the pose error is reduced to the specified threshold or the maximum number of iterations is reached. The final arm configuration, represented by link model parameters, is then obtained. Through a reverse transformation by LINK2ARC, the bending and rotation angles of the three-segment continuum robot are determined, completing the inverse kinematics solution.

\begin{table*}[!h]
\centering
\caption{The simulation result of solving inverse kinematic under different methods for continuum robot.}
\footnotesize
\label{Table2}
\begin{threeparttable}
\resizebox{\textwidth}{!}{ 
\begin{tabular}{ccccccccc}
\toprule
\multirow{2}{*} {\textbf{Segments}} & \multirow{2}{*}{\textbf{Methods}} & \multirow{2}{*} {\makecell{\textbf{Success}\\ \textbf{Rate}}} & \multicolumn{3}{c}{\textbf{Number of Average Iteration}} & \multicolumn{3}{c}{\textbf{Average Time Cost/ms}} \\ 
 &  &  & \textbf{Top 20\%} & \textbf{Top 60\%} & \textbf{Top 100\%} & \textbf{Top 20\%} & \textbf{Top 60\%} & \textbf{Top 100\%} \\ 
\midrule
2 & TLGI & 100\% & 1.61 & 1.87 & 2.08 & 0.33 & 0.45 & 0.66 \\ 
2 & TL-FABRIKc & 100\% & 1.61 & 1.87 & 2.08 & 0.35 & 0.63 & 0.84 \\ 
2 & TLGI*\tnote{1} & 100\% & 1.61 & 1.87 & 2.08 & 0.33 & 0.51 & 0.74 \\ 
2 & TL-FABRIKc*\tnote{2} & 100\% & 1.61 & 1.87 & 2.08 & \textbf{0.31} & \textbf{0.44} & \textbf{0.64} \\ 
3 & TLGI & 83.1\% & \textbf{6.90} & 23.66 & 415.87 & 3.02 & 10.29 & 181.08 \\ 
3 & TL-FABRIKc & \textbf{92.6\%} & 7.78 & 27.09 & 328.97 & 3.40 & 10.36 & 115.83 \\ 
3 & TLGI* & 80.7\% & 7.35 & \textbf{20.94} & 442.11 & 3.20 & \textbf{9.11} & 191.48 \\ 
3 & TL-FABRIKc* & 92.1\% & 6.91 & 24.01 & \textbf{307.75} & \textbf{2.99} & 10.12 & \textbf{115.57} \\ 
4 & TLGI & 88.8\% & \textbf{5.09} & 11.28 & 265.04 & \textbf{2.24} & 4.94 & 116.02 \\
4 & TL-FABRIKc & \textbf{95.0\%} & 5.42 & 12.90 & \textbf{180.39} & 2.37 & 5.32 & \textbf{64.18} \\ 
4 & TLGI* & 86.4\% & 5.28 & \textbf{10.11} & 296.80 & 2.34 & \textbf{4.47} & 131.27 \\ 
4 & TL-FABRIKc* & 93.9\% & \textbf{5.09} & 11.16 & 198.86 & \textbf{2.24} & 4.84 & 76.56 \\ 
8 & TLGI & 75.6\% & \textbf{5.63} & 14.68 & 514.87 & \textbf{2.67} & 6.93 & 242.92 \\
8 & TL-FABRIKc & \textbf{95.8\%} & 7.17 & 17.38 & \textbf{176.73} & 3.37 & 7.58 & \textbf{69.27} \\ 
8 & TLGI* & 76.2\% & 6.63 & \textbf{14.00} & 493.70 & 3.15 & \textbf{6.63} & 233.60 \\
8 & TL-FABRIKc* & 90.0\% & 5.66 & 14.88 & 282.26 & \textbf{2.67} & 6.79 & 122.03 \\ 
 \bottomrule
\end{tabular}
}
\begin{tablenotes}    
        \footnotesize               
        \item[1] TLGI* denotes TLGI with the improvement of \textit{Problem} 2.
        \item[2] TL-FABRIKc* denotes TL-FABRIKc without the improvement of \textit{Problem} 2.
\end{tablenotes}            
\end{threeparttable}
\end{table*}

\subsection{Simulation Research}
\label{Section3.3}

To validate the effectiveness of the proposed TL-FABRIKc algorithm in solving the inverse kinematics of multi-segment continuum robots, simulations and comparative analyses were conducted using TLGI algorithm, TL-FABRIKc algorithm, TLGI* algorithm and TL-FABRIKc* algorithm on two-, three-, four-, and eight-segment continuum robots under six degree of freedom constraints. In the simulations, the bending angle and rotation angle ranged from $[0,0.5\pi ]$ and $[0,2\pi ]$, respectively. Each scenario included 5,000 inverse kinematics tasks. The initial arm configuration parameters for each task were generated randomly within specified ranges, and the target poses were derived from another set of random arm configurations using the forward kinematics model in Equation (2). Thus, both the initial configurations and target poses for each task were randomized. For the six degree of freedom constraints, the maximum allowable end-effector position error was set to 0.01 mm, with a rotational deviation of no more than 0.2° and maximum iteration times for 2000. The simulations were performed on a hardware platform running Windows 10, powered by an Intel(R) Core(TM) i7-9700 CPU @3.00GHz, using MATLAB 2023a as the software platform.

\begin{figure*}[h]
\centering
\includegraphics[width=4.7in]{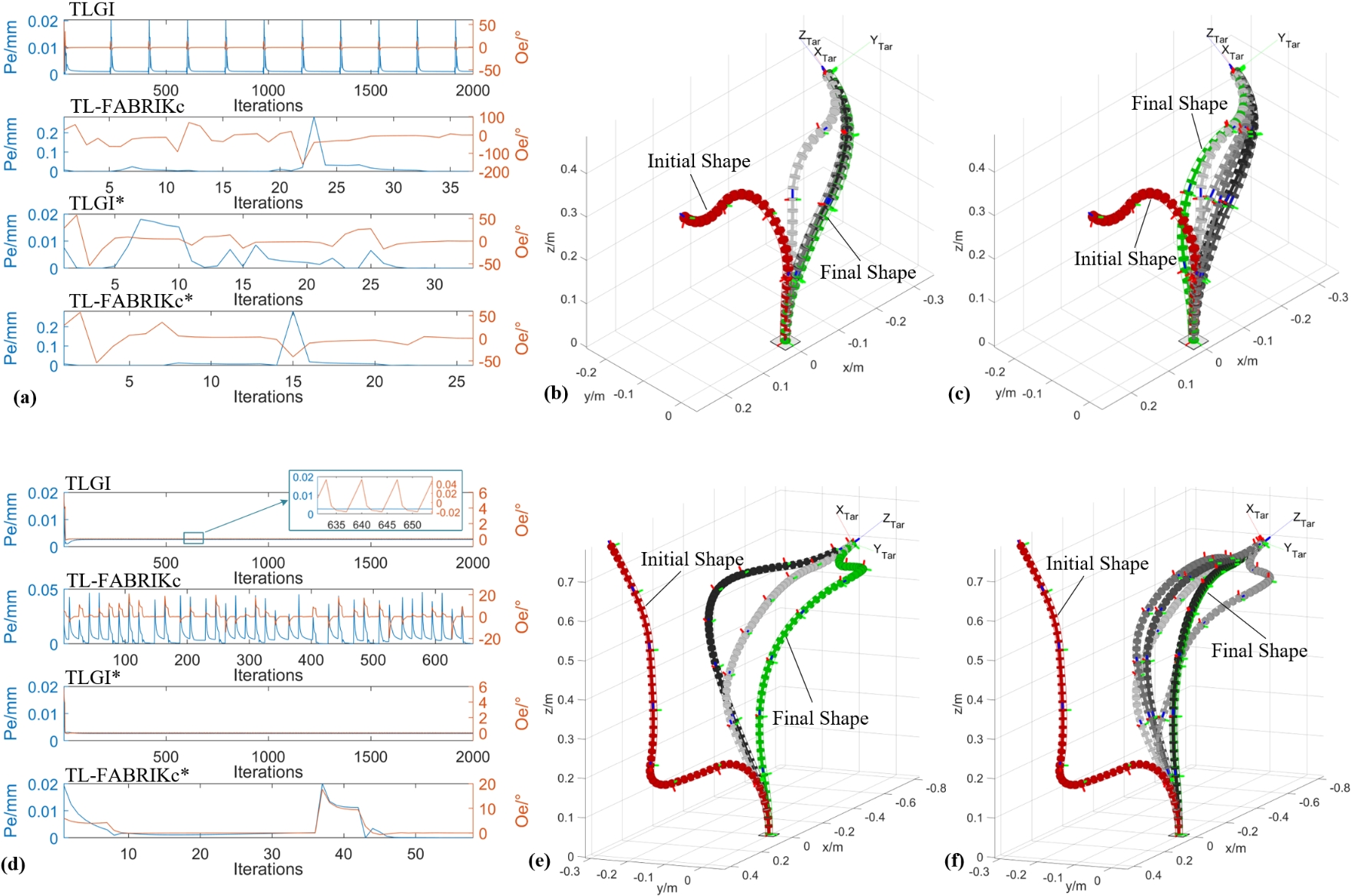}
\caption{The iteration processes of solving inverse kinematic under TL-FABRIKc and TLGI for continuum robot. (a) Error variation in an iteration for 4-segment. (b) Shape variation in TLGI. (c) Shape variation in TL-FABRIKc. (d) Error variation in an iteration for 8-segment. (e) Shape variation in TLGI. (f) Shape variation in TL-FABRIKc.}
\label{figure6}
\end{figure*}

The simulation results under these conditions are summarized in Table \ref{Table2}. The TL-FABRIKc algorithm consistently achieved the highest success rates across all four configurations of continuum robots. For the two-segment continuum robot, which lacks kinematic redundancy, all four geometric iterative strategies completed the task successfully, with an average of only two iterations required per solution and an average computation time of just 0.5 ms. Given the extensive research and application of two-segment continuum robots, TL-FABRIKc represents a valuable choice for solving inverse kinematics problems in experimental platforms for this type of robot. For the eight-segment continuum robot, the TL-FABRIKc method achieved an overall success rate of 95.8\%, which is 20\% higher than the TLGI method. Which means more than 1,000 additional successful task solutions  were obtained within the expected error range when using the proposed method than TLGI.

In terms of average iteration count, Table \ref{Table2} ranks the number of iterations required for each task from least to most and calculates the average iterations at the top 20\%, 60\%, and 100\%. The iteration count corresponds to the total number of complete inner loop iterations executed. For the three-segment scenario, the TL-FABRIKc* algorithm achieved the lowest overall average iteration count of 307.75, indicating that the improvement of \textit{Problem} 1 significantly reduced the number of iterations needed for a solution. In the four-segment and eight-segment scenarios, the TL-FABRIKc algorithm reduced the average number of iterations by 84.65 and 338.14, respectively, compared to the TLGI method. For the top 20\% of tasks, although the TLGI method required fewer iterations than other strategies, the iteration counts for TLGI and TL-FABRIKc* were nearly identical, with other methods requiring slightly more iterations. This suggests that for the top 20\% of tasks, the difficulty level was similar across all four methods. However, the data for the 60th and 100th percentiles reveal that TLGI’s slight advantage diminishes. This also confirms that the problems discussed in Section \ref{Section3.1}, namely \textit{Problem} 1 and 2 are prevalent. Although the maximum allowed iteration count of 2000 enabled TLGI to solve some difficult tasks, it came at the cost of longer computation times and a higher number of iterations. The inability to solve within the limited iteration count directly led to the lower overall success rate for TLGI compared to the other methods.

Regarding computation time, the total solution times were similarly ranked, and the average times were calculated for the top 20\%, 60\%, and 100\%. Similar to the iteration counts, the fastest average solution times for the top 20\% and 60\% of tasks were close across different methods in the three-, four-, and eight-segment scenarios. However, the TL-FABRIKc method had the shortest overall average solution time, with the eight-segment scenario taking only 68.27 ms, which is 173.65 ms less than the TLGI method.

Furthermore, Fig.\ref{figure6} illustrates the iteration process for solving a task with four-segment and eight-segment continuum robots. As shown in Fig.\ref{figure6}(a), the TLGI method failed to converge within 2000 iterations, with both the end-effector position and orientation exhibiting periodic oscillations. This reflects the limitations described in \textit{Problem} 1 and 2. However, TL-FABRIKc, TLGI*, and TL-FABRIKc* were able to avoid oscillations due to the incorporation of the improvements for \textit{Problem} 1 or 2. Fig.\ref{figure6}(b) and 6(c) display six selected configurations during the iteration process. The red indicates the initial configuration, green represents the final configuration after the last iteration, and varying shades of gray show intermediate configurations. The repeated oscillations inherent in the TLGI method resulted in multiple intermediate configurations overlapping. In contrast, the TL-FABRIKc method clearly shows distinct intermediate configurations during the iteration process.

Fig.\ref{figure6}(d)–(f) show the solution process for the eight-segment continuum robot. Both TLGI and TLGI* experienced oscillations in the end-effector orientation during iterations, ultimately leading to solution failure. Although the TL-FABRIKc method also exhibited oscillations during the iteration process, this was primarily due to the method entering workmode 1–3. However, the presence of workmode 4 prevented the oscillations from becoming strongly periodic, allowing the solution to converge successfully after 650 iterations. TL-FABRIKc*, despite lacking the improvement for \textit{Problem} 2, completed the solution process in just 59 iterations due to the efficiency of workmode 4. The comparison in Fig.\ref{figure6}(e) and (f) highlights how the TL-FABRIKc algorithm, with improvements for \textit{Problem} 1 and 2, overcomes the local oscillation problems caused by workmode 1–3, and successfully solves the inverse kinematics of the continuum robot with the required precision. The 3D configurations of the continuum robot depicted in the figures were rendered using the CRVisToolkit toolbox \cite{bib13}.

\section{Solver for Motion Planning}
\label{Chapter4}
\subsection{TL-FABRIKc Under Environment Constraint}
\label{Section4.1}

In solving inverse kinematics problems based on geometric iterative strategies, the spatial position updates of each virtual joints determine the shape and spatial coverage of each segment. However, all methods that update virtual joint positions based on the original FABRIK strategy fail to eliminate the influence of the initial configuration on the virtual joint position updates. This, in turn, leads to the problem where virtual joints only have a single candidate position, constrained by the initial configuration. Therefore, under the influence of constraints, the strategy for updating virtual joint positions needs to be adjusted. 

\begin{figure*}[!t] 
\begin{equation}
\label{1-13}
\left\{ \begin{array}{c}
\begin{aligned}
\overrightarrow{{{v}_{c}}} &=\frac{[-\overrightarrow{vl_{je}^{*}}(2),\overrightarrow{vl_{je}^{*}}(1),0]}{\sqrt{\overrightarrow{vl_{je}^{*}}{{(2)}^{2}}+\overrightarrow{vl_{je}^{*}}{{(1)}^{2}}}},\text{ } &&if\text{ }\overrightarrow{vl_{je}^{*}}(1)\ne 0\text{ or }\overrightarrow{vl_{je}^{*}}(2)\ne 0 \\ 
\overrightarrow{{{v}_{c}}} &=[1,0,0],\text{ } &&if\text{ }\overrightarrow{vl_{je}^{*}}(1)=0\text{ and }\overrightarrow{vl_{je}^{*}}(2)=0 \\ 
\end{aligned}
\end{array} \right.
\end{equation}
\end{figure*}

\begin{figure}[!t]
\centering
\includegraphics[width=3.3in]{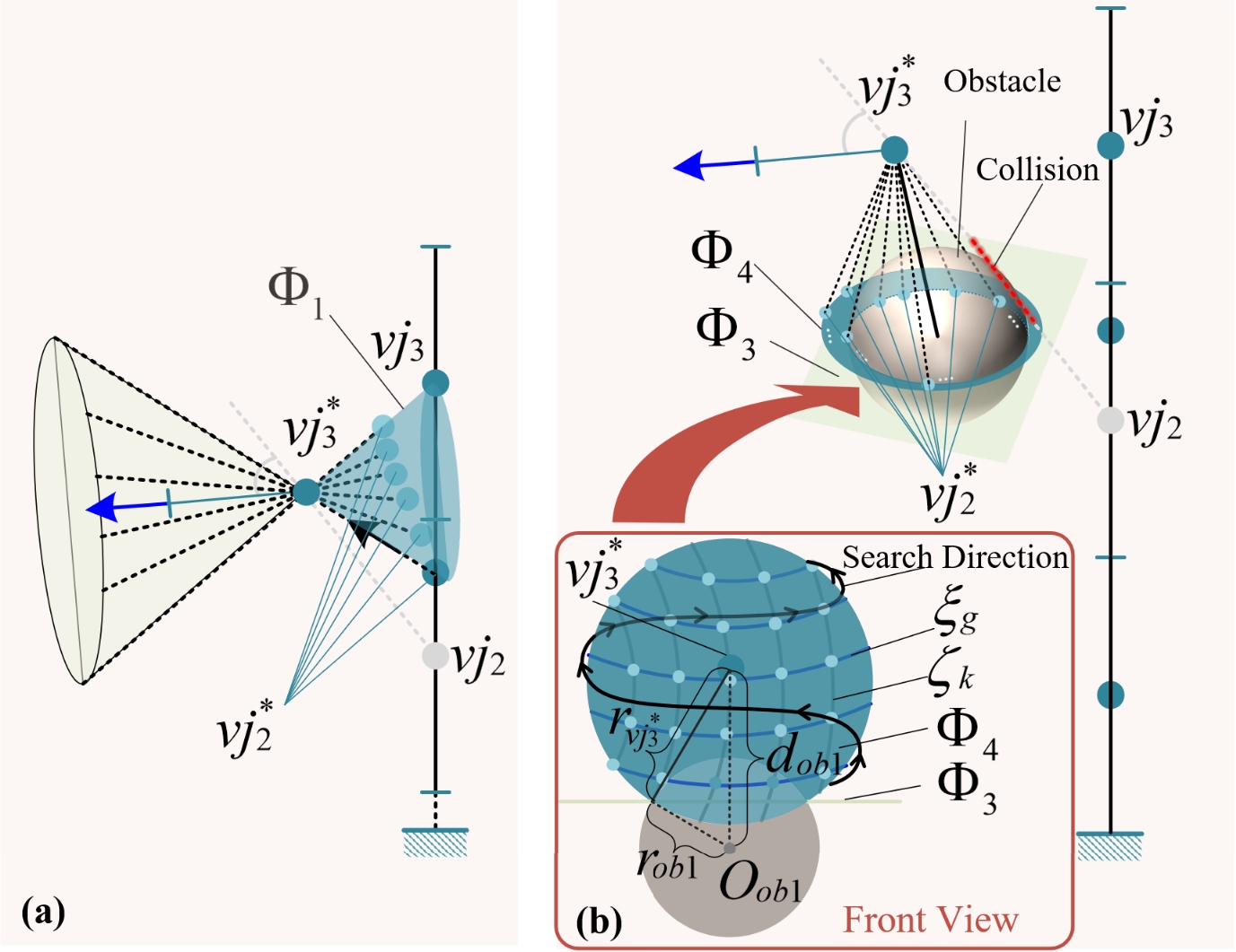}
\caption{Joint and obstacle constraint strategies of TL-FABRIKc. (a) Joint constraint strategy of TL-FABRIKc. (b) Obstacle constraint strategy of TL-FABRIKc.}
\label{figure7}
\end{figure}

To increase the number of candidate points for the virtual joint and achieve more configuration solutions under the bending angle constraints, the TL-FABRIKc method introduces a conical surface update strategy, as shown in Fig.\ref{figure7}(a). Here, the candidate point $vj_{2}^{*}$ lies on the inverse conical surface ${{\Phi }_{1}}$, which is determined by the maximum bending angle and the direction of the end-effector. The calculation of the candidate point is given by Equation (14),(15) and (16). In this context, $\overrightarrow{vl_{je}^{*}}$ represents the unit vector of the current segment end direction, with the index in parentheses indicating a specific element in the vector. The vector $\overrightarrow{vl_{je}^{*}}$ is considered the central axis of the conical surface ${{\Phi }_{1}}$, and a perpendicular vector $\overrightarrow{{{v}_{c}}}$, is determined as shown in Equation (14). A random rotation angle ${{\theta }_{r}}$ is chosen, and the vector $\overrightarrow{{{v}_{c}}}$ is rotated about the axis $\overrightarrow{vl_{je}^{*}}$ to produce an arbitrary random vector $\overrightarrow{{{v}_{cr}}}$ on the conical surface while maintaining its perpendicularity to the central axis. The $\text{Rot}(\cdot )$ function in Equation (15) refers to the rotation function described in Equation (11). Under the constraint of the maximum bending angle $\theta _{3}^{**}$, a specific vector $b$ on the conical surface is determined, as shown in Equation (16). The virtual joint $vj_{2}^{*}$ is then located on this vector according to the length of the virtual link for the segment.

\begin{equation}
\label{1-14}
\overrightarrow{{{v}_{cr}}}=\text{Rot}(\overrightarrow{vl_{je}^{*}},{{\theta }_{r}})\cdot \overrightarrow{{{v}_{c}}}.
\end{equation}

\begin{equation}
\label{1-15}
b=\overrightarrow{{{v}_{cr}}}\sin \theta _{3}^{**}+\overrightarrow{vl_{je}^{*}}\cos \theta _{3}^{**}.
\end{equation}

Correspondingly, as shown in Fig.\ref{figure7}(b), the TL-FABRIKc method constructs a spherical surface ${{\Phi }_{4}}$ centered at $vj_{3}^{*}$ with a radius ${{r}_{vj_{3}^{*}}}$, where ${{r}_{vj_{3}^{*}}}$ is calculated based on the size and position of obstacles. On this spherical surface, there are countless candidate points for $vj_{2}^{*}$. To facilitate iterative calculation in the TL-FABRIKc algorithm, Fig.\ref{figure7}(b) introduces a meridian and parallel search method to find feasible virtual joint points that avoid collisions within the ${{\Phi }_{4}}$ region. Before starting the search, the starting point and direction along the meridian are determined. The search begins at the candidate point indicated by the normal vector of the plane intersecting the obstacle sphere and passing through $vj_{3}^{*}$, first searching towards the lower latitude region and then towards the higher latitude region. The search direction along the parallel is as indicated by the spiral arrow in Fig.\ref{figure7}(b).

\subsection{Motion Planning Algorithms for the Case of a Floating Base}
\label{Section4.2}
The workspace and mobility of continuum robots with a fixed base are inherently limited. To overcome these constraints, many research teams have mounted continuum robots on various floating bases, aiming to adapt to more complex environments. The iterative principle of the TL-FABRIKc method ensures that the end orientation remains aligned with the target orientation across successive iterations. Additionally, the underlying FABRIKc iterative strategy maintains a strong spatial correlation between the arm configurations of consecutive iterations. Follow-the-leader strategies are widely used in continuum robots with mobile bases and in robots with high degrees of redundancy. In continuum robots, the follow-the-leader strategy works by ensuring that the new trajectory to be followed largely coincides with the arm configuration formed in the previous iteration. Utilizing these two properties, the TL-FABRIKc-FTL algorithm, which takes into account the base floating requirement, is proposed based on the TL-FABRIKc algorithm by combining the joint and obstacle constraint strategies.

\begin{figure}[!t]
\centering
\includegraphics[width=3.4in]{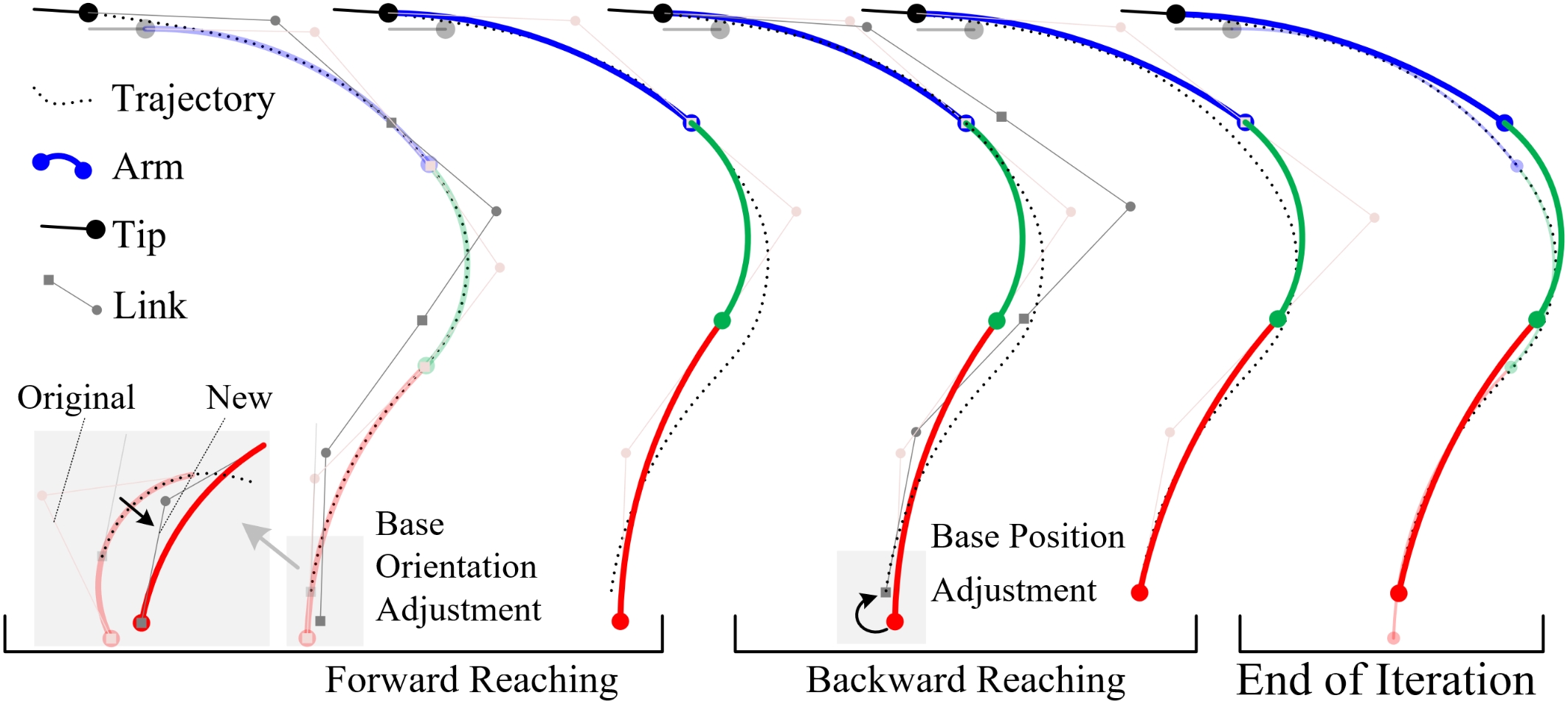}
\caption{The strategies of TL-FABRIKc for follow-the-leader task.}
\label{figure8}
\end{figure}

As shown in Fig.\ref{figure8}, the thick solid lines with red, green, and blue double-arrow circles represent the current arm configuration, while the black dashed lines indicate the target position for the next iteration. Based on the data represented by the black dashed lines, the ARC2LINK function in Algorithm \ref{Algorithm1} is used to derive the FABRIKc parameters for the target arm configuration. These parameters are updated before each follow-the-leader iteration begins and are used as the initial configuration for that iteration. The strategy for follow-the-leader task first completes the forward iteration phase as described in Fig.\ref{figure5}(b), where the base orientation is adjusted from its original direction to the new direction specified by the trajectory, as shown on the left side of Fig.\ref{figure8}. Next, the algorithm completes the backward iteration phase, illustrated in Fig.\ref{figure5}(c), where, due to the floating base, the base position is adjusted to the starting point of the trajectory before the backward iteration begins, as depicted in the center of Fig.\ref{figure8}. After several iterations, the new arm configuration that meets the required conditions is achieved, as shown on the right side of Fig.\ref{figure8}.

\begin{algorithm}[H]
\caption{Strategy for Follow-the-leader Task Under Environmental Constraints based on TL-FABRIKc. (TL-FABRIKc-FTL)}\label{Algorithm2}
\begin{algorithmic}[1]
\Require{Tip Position of $j^{\text{th}}$ Segment $O_{je}^{*}$; Virtual Joint of $j^{\text{th}}$ Segment $vj_{j}^{*}$; Old Virtual Joint of $j-1^{\text{th}}$ Segment $v{{j}_{j-1}}$; Position and Size of $n^{\text{th}}$ Obstacle $O_{ob}^{n}$, $r_{ob}^{n}$; Number of Obstacles $N$; Max bend angle of $j^{\text{th}}$ Segment $\theta _{\max }^{j}$; Trajectory to be tracked $\vartheta$; Initial Shape $\theta_{i}^{i}, \varphi_{i}^{i}$; Segment Length ${{L}_{i}}$}
\Ensure{Virtual Joint of $j-1^{\text{th}}$ Segment $vj_{j-1}^{*}$}

\State $(\theta_{i}^{i}, \varphi_{i}^{i}) = \Call{TRA\_ARC2LINK}{\theta _{i}^{i},\varphi _{i}^{i},{{L}_{i}},\vartheta}$
\If{$\exists  n\in [1,N]$, $\Call{ADCEK}{vj_{j}^{*}, v{{j}_{j-1}}, O_{ob}^{n}, r_{ob}^{n}, \theta _{\max }^{j}} == \textbf{false}$}
    \State $(vj_{j-1}^{{{\Phi }_{1}}}) = \Call{THETAG}{O_{je}^{*}, vj_{j}^{*}, \theta _{\max }^{j}, \theta_{i}^{i}, \varphi_{i}^{i}}$
    \State $(vj_{j-1}^{{{\Phi }_{4}}}) = \Call{OBSTAG}{vj_{j}^{*}, O_{ob}^{n}, r_{ob}^{n}, \theta_{i}^{i}, \varphi_{i}^{i}}$
    \For{$i=1$ to ${{N}_{\zeta }}$}
        \State $({{\zeta }_{i}}) = \Call{FINDLALI}{vj_{j-1}^{{{\Phi }_{4}}}, i}$
        \For{$ii=1$ to ${{N}_{\xi }}$}
            \State $({{\xi }_{ii}}) = \Call{FINDLOLI}{vj_{j-1}^{{{\Phi }_{4}}}, ii}$
            \State $(vj_{j-1}^{\diamond }) = \Call{LALOLI2VJ}{{{\zeta }_{i}}, {{\xi }_{ii}}, vj_{j-1}^{{{\Phi }_{1}}}}$
            \If{$\forall n\in [1,N]$, $\Call{ADCEK}{vj_{j}^{*}, vj_{j-1}^{\diamond }, O_{ob}^{n}, r_{ob}^{n}, \theta _{\max }^{j}} == \textbf{true}$}
                \State $vj_{j-1}^{*} = vj_{j-1}^{\diamond }$
                \State \textbf{break}
            \EndIf
        \EndFor
    \EndFor
    \If{$vj_{j-1}^{*} == \textbf{nil}$}
        \State $vj_{j-1}^{*} = v{{j}_{j-1}}$
    \EndIf
\Else
    \State $vj_{j-1}^{*} = v{{j}_{j-1}}$
\EndIf
\State \textbf{return} $vj_{j-1}^{*}$
\end{algorithmic}
\end{algorithm}

\begin{figure*}[!b]
\centering
\includegraphics[width=5in]{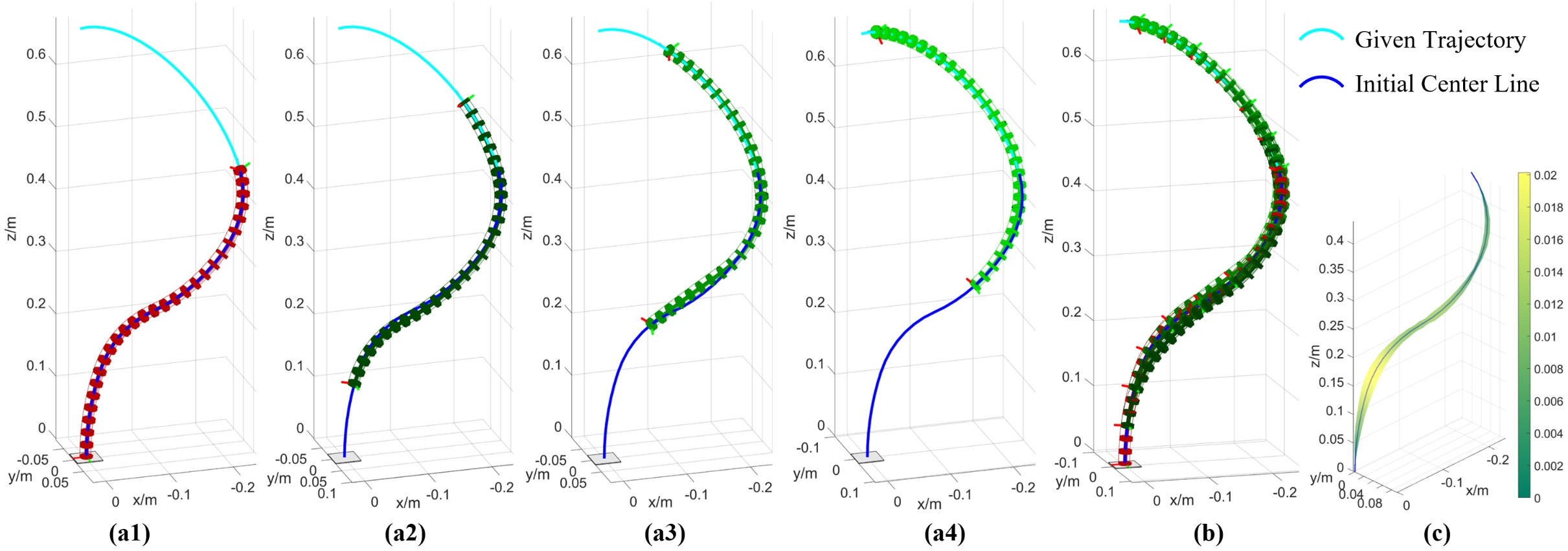}
\caption{The simulation result of TL-FABRIKc-FTL algorithm. (a) The motion sequence in follow-the-leader simulation. (b) The motion sequence for follow-the-leader simulation task in one sub-figure. (c) Average error band.}
\label{figure9}
\end{figure*}

Algorithm \ref{Algorithm2} gives the pseudo-code of the TL-FABRIKc-FTL algorithm for solving the follow-the-leader task in the case of TL-FABRIKc using the new virtual joint adjustment strategy. Taking the virtual joint $vj_{j-1}^{*}$ for segment $j-1$ as an example, based on the FR and BR processes described in Fig.\ref{figure5}(b1)-(b2) and (c1)-(c2), the parameter iteration for segment $j$ begins, determining the virtual joint $vj_{j}^{*}$. TRA\_ARC2LINK is updated from the ARC2LINK function in Algorithm \ref{Algorithm1} and is used to associate the information of the trajectory $\vartheta$ to be tracked with the initial arm shape in order to obtain the new arm shape parameters $\theta_{i}^{i}, \varphi_{i}^{i}$. The ADCEK function checks the original virtual joint $v{{j}_{j-1}}$for compliance with joint or obstacle constraints. If any non-compliance is detected, subsequent calculations are performed. First, the cone surface ${{\Phi }_{1}}$ satisfying the bending angle constraint $\theta _{\max }^{j}$and the candidate point $vj_{j-1}^{{{\Phi }_{1}}}$ are determined by the THETAG function in the pseudocode. Similarly, the spherical surface ${{\Phi }_{4}}$ and the candidate point $vj_{j-1}^{{{\Phi }_{4}}}$ are determined by the OBSTAG function based on the obstacle information. The candidate points on the two surfaces are utilized to search and rank candidate points in the direction shown in Fig. \ref{figure7}(d), i.e., the functions FINDLALI and FINDLOLI functions, and transformed by the function LALOLI2VJ to obtain the candidate point $vj_{j-1}^{\diamond}$ to be calibrated, and finally calibrated again to check that ADCEK complies with all the obstacles and the bending angle constraints, and ultimately obtains the virtual joint $vj_{j-1}^{*}$ that meets the requirements.

\subsection{Simulation Research}
\label{Section4.3}

To validate the performance of the TL-FABRIKc-FTL algorithm in follow-the-leader tasks, an extended trajectory was set at the end of an initial arm configuration, as shown by the cyan solid line in Fig.\ref{figure9}(a1). The parameters of the initial arm shape and the trajectory are given in Equation (17), where ${{r}_{tra}}$ represents the radius of the arc on which the trajectory lies, ${{l}_{tra}}$ indicates the trajectory length. In accordance with the characteristics of a follow-the-leader task, the trajectory that the continuum robot needs to follow is composed of both the original arm centerline and the newly extended trajectory at the end, where the blue solid line represents the original arm centerline. Fig.\ref{figure9}(a1)-(a4) demonstrate how the current arm configuration approaches the target trajectory under different base positions. In Fig.\ref{figure9}(b), seven sets of tracking arm configurations are plotted at intervals of 60 mm along the tracking direction. The evident overlap in arm positions indicates a good overall tracking performance. Furthermore, Fig.\ref{figure9}(c) was plotted to analyze the relationship between the maximum tracking error and the position on the arm during the tracking process. From the color-coded error map, it can be observed that the continuum robot achieves the best proximity between the end and the base during the tracking of the specified trajectory. The maximum tracking error of 20 mm occurs at the junction between the first and second segments, with an average tracking error of 4.06 mm.

\begin{equation}
\label{1-13}
\left\{ \begin{array}{c}
\begin{aligned}
  & {\theta}_{init} =[0.29,0.77,0.70,1.01] \\ 
 & {\varphi}_{init} =[2.75,0.40,4.81,4.99] \\ 
 & {{r}_{tra}}=200mm,{{l}_{tra}}=400mm \\ 
\end{aligned}
\end{array} \right.
\end{equation}

\section{Experiment}
\label{Chapter5}
\begin{figure}[!t]
\centering
\includegraphics[width=3.1in]{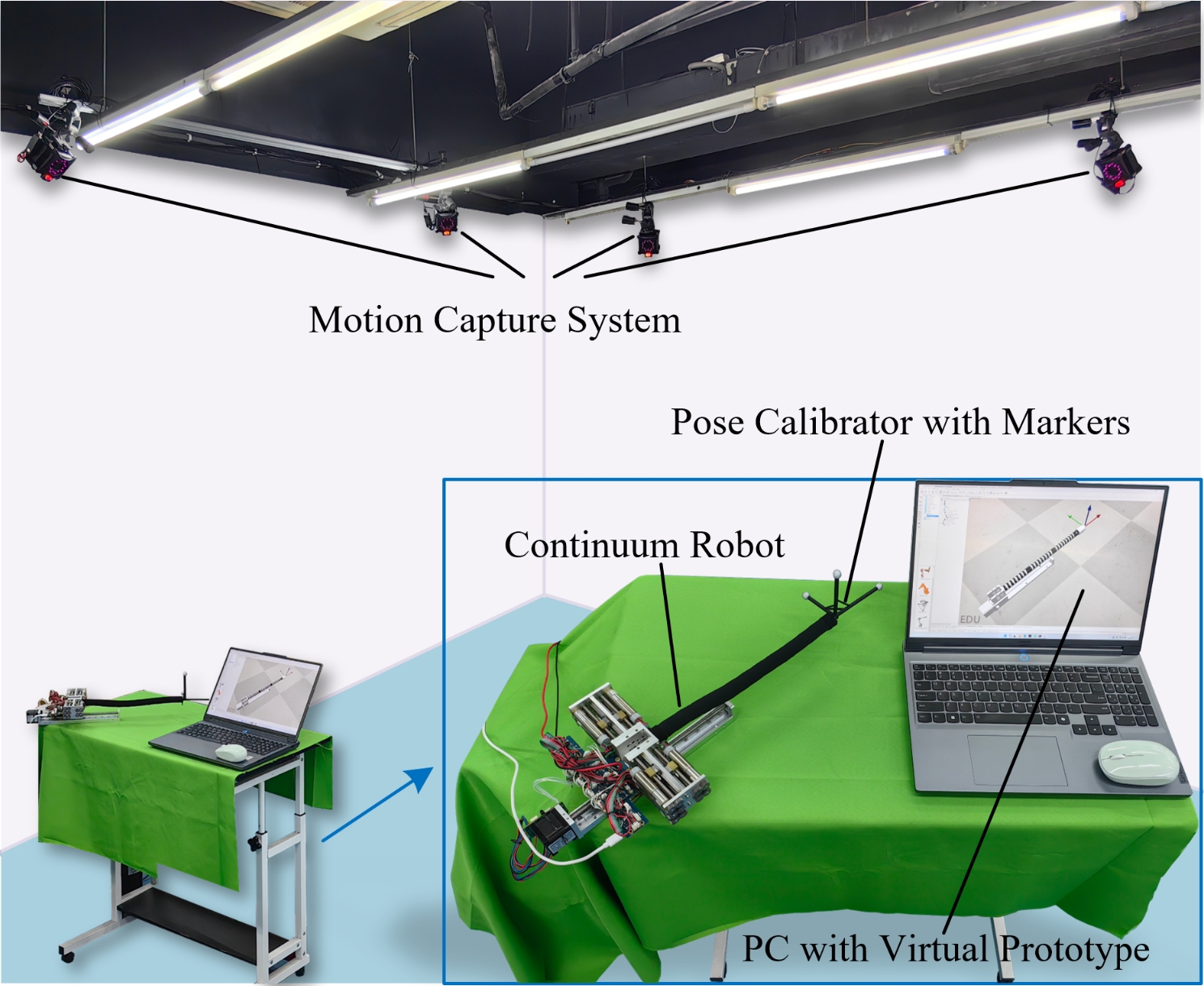}
\caption{Three-segment continuum robot with moving base.}
\label{figure12}
\end{figure}

To validate the effectiveness of the TL-FABRIKc algorithm and its extensions, an experimental platform for a three-segment continuum robot with a telescopic base was built, as shown in Fig.\ref{figure12}. The platform consists of the robot prototype system, an optical tracking system, and a computer. To accurately capture the end-effector pose, a pose calibration device consisting of multiple rods and reflective spheres was mounted at the end. There are three individual experiments conducted for different proposes. Robot motion accuracy calibration, end-point tracking under a fixed base, and follow-the-leader task under a floating base are included. During the experiments, the control variable for manipulating the robot configuration was the change in the length of the driving wires. After calculating the arm configuration, the TL-FABRIKc algorithm derived the control variable using Equation (8).

To enhance the visualization of the experimental process, a scaled high-fidelity virtual prototype of the continuum robot was developed on the CoppeliaSim simulation platform \cite{bib32}. The virtual prototype integrates the TL-FABRIKc algorithm and its extensions. Real-time synchronization between the physical and the virtual prototypes is achieved by transmitting the control variable data, specifically the cable length variations, through serial communication. In order to fully illustrate the effectiveness of the TL-FABRIKc algorithm, the experiments carried out in this paper use an open-loop control strategy.

\subsection{Experiment 1: Robot Motion Accuracy Calibration}
\label{Section5.1}

The purpose of the accuracy calibration experiment is to quantify the positional error between the physical continuum robot prototype and the virtual one under the action of the same signal. This error mainly stems from the static friction between tendons and inner hole and deviations in the constant curvature model. In the experiment, bending and rotational motions were performed on the second and last segments, respectively, and the positional data from both the virtual and physical prototypes were recorded. The bending angle of the second segment varied from 0° to 60°, and the bending angle of the end segment varied from 0° to 90°. The rotation angle for both segments varied from 0° to 360°. Since the goal of \textit{Experiment} 1 was to determine the positional accuracy of the physical prototype, the virtual prototype data were used as the ideal reference for direct comparison with the physical prototype data.

\begin{figure}[!t]
\centering
\includegraphics[width=3.5in]{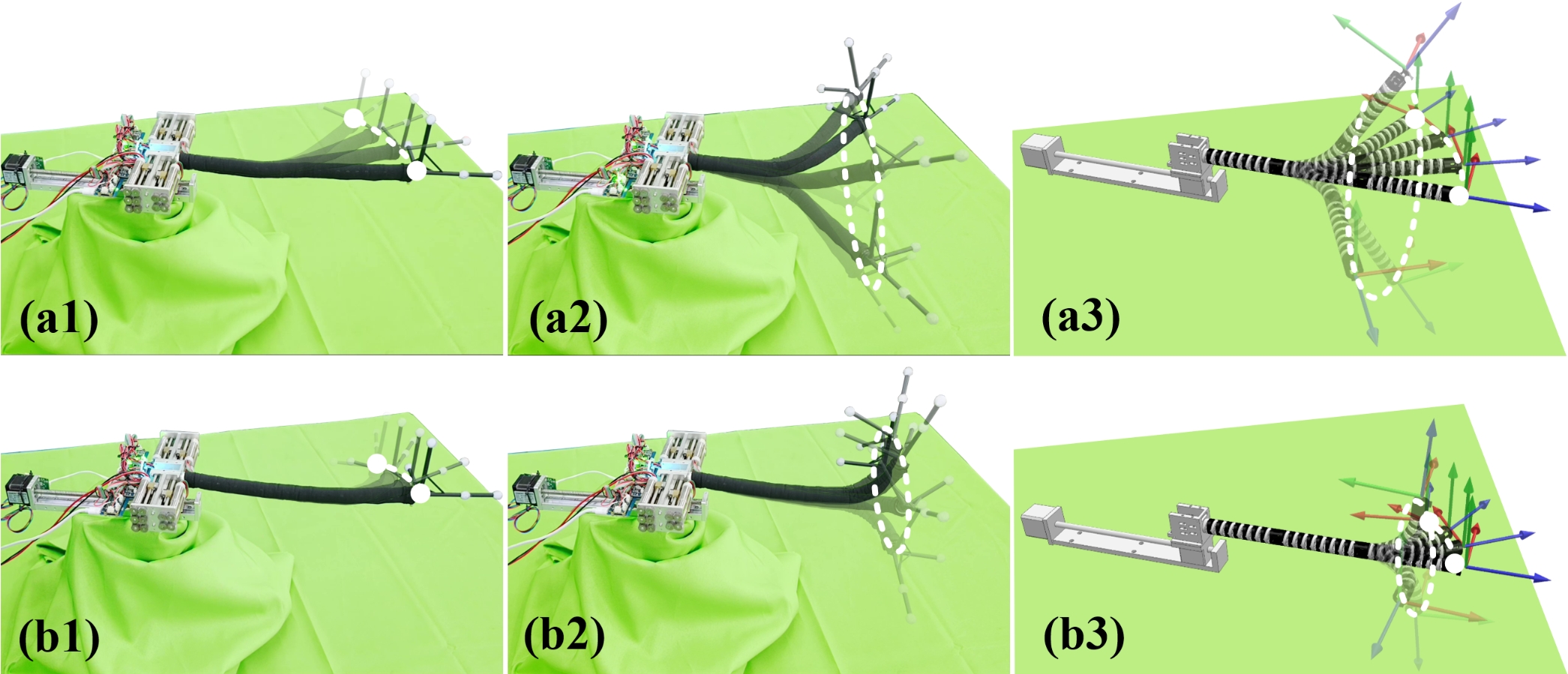}
\caption{Motion sequence in \textit{Experiment} 1. (a1)-(a3) Physical experiments in bending, rotating and simulation in CoppeliaSim for $2^{\text{th}}$ segment. (b1)-(b3) Physical experiments  in bending, rotating and simulation in CoppeliaSim for $3^{\text{th}}$ segment. }
\label{figure13}
\end{figure}

\begin{figure}[!t]
\centering
\includegraphics[width=3.2in]{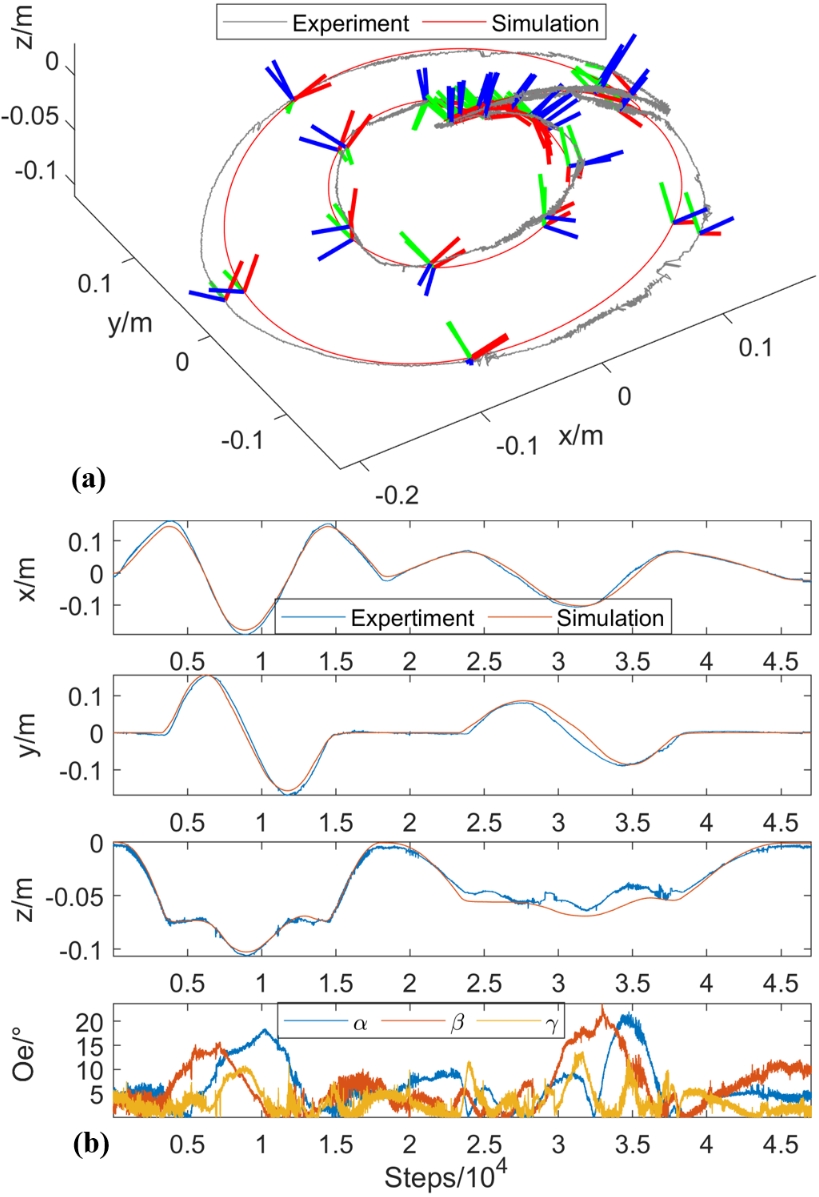}
\caption{Tip pose data analysis in \textit{Experiment} 1. (a) Tip trajectory. (b) Pose error.}
\label{figure14}
\end{figure}

Fig.\ref{figure13} records the trajectory sequences of the second and third segment for both the physical and virtual prototypes during motion. In the physical prototype motion tests, the second segment and the end segment exhibited independent movement, with no significant interference between segments. Fig.\ref{figure14} illustrates the changes in the end-effector pose for both prototypes in \textit{Experiment} 1. The average positional deviations in the $x,y,z$ directions were 6.8 mm, 7.4 mm, and 4.2 mm, respectively, with an average distance deviation of 12.7 mm. The average orientation deviations for the end-effector in the $\alpha ,\beta ,\gamma $ directions were 6.37°, 6.41°, and 5.43°, respectively. These accuracy data serve as a benchmark for error analysis in future prototype experiments.

\subsection{Experiment 2: End-Effector Point Tracking with a Fixed Base}
\label{Section5.2}

\begin{figure}[!t]
\centering
\includegraphics[width=3.5in]{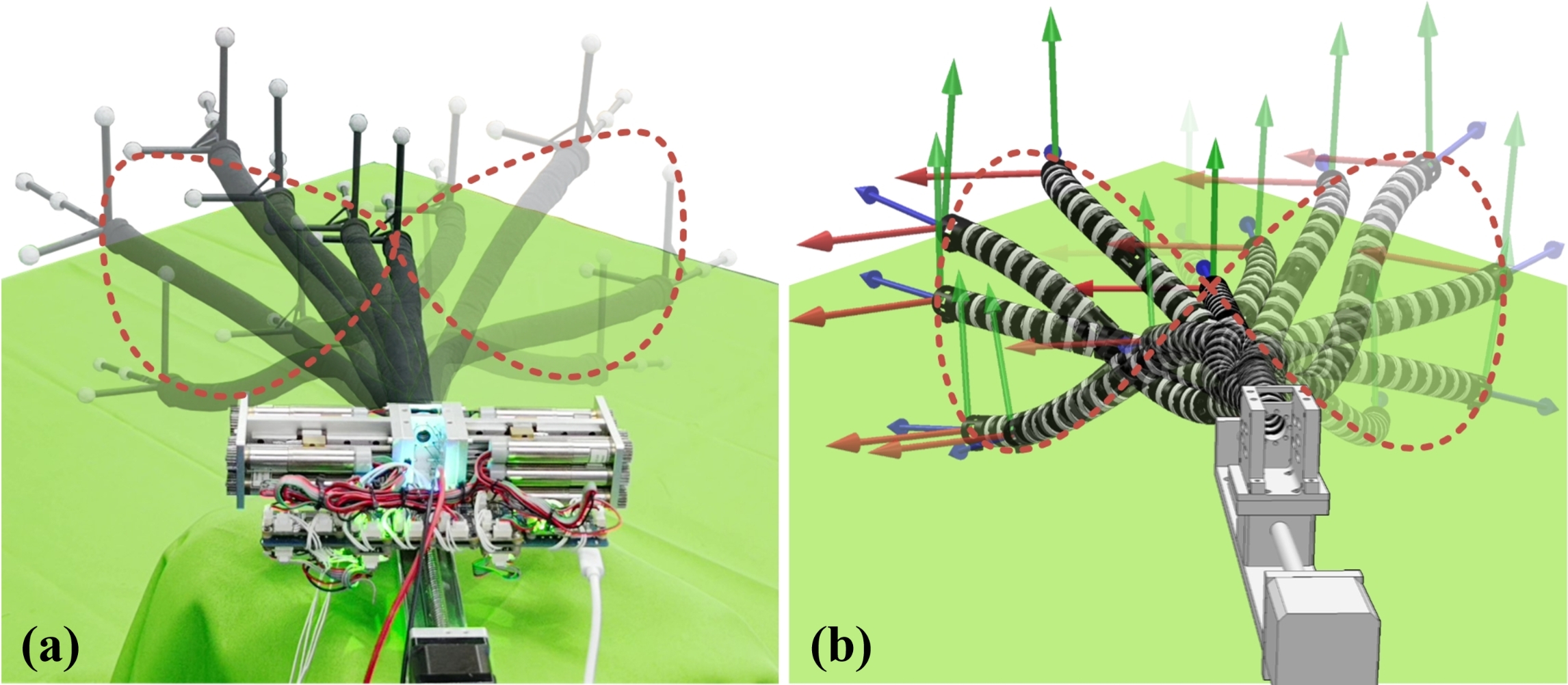}
\caption{Motion sequence in \textit{Experiment} 2. (a) '$\infty$' trajectory tracking experiment. (b) '$\infty$' trajectory tracking simulation in CoppeliaSim.}
\label{figure15}
\end{figure}

The objective of \textit{Experiment} 2 is to verify the accuracy and performance of the TL-FABRIKc algorithm in solving the specified end-effector poses when the robot base is fixed. For this purpose, a 3D '$\infty $' trajectory was designed, with its fundamental equation shown in Equation (18), where ${{\theta }_{rot}}$ represents the angle by which the left and right halves of the original trajectory are symmetrically rotated towards the center axis. As shown in Fig.\ref{figure15}, the robot prototype and virtual one coordinated across all segments to track the trajectory, with both showing consistent changes in end-effector orientation throughout the process. Based on the experimental data analysis, the physical prototype exhibited average positional deviations of 8.3 mm, 9.8 mm, and 6.9 mm in $x,y,z$ directions, with a distance deviation of 16.7 mm. In contrast, the virtual prototype showed much smaller deviations of 1.2 mm, 1.4 mm, 0.6 mm, and 2.1 mm, respectively. For the end-effector orientation, the physical prototype had average deviations of 5.53°, 6.12°, and 5.7° in the $\alpha ,\beta ,\gamma $ axes, while the virtual prototype deviations were 0.54°, 0.59°, and 0.13°, respectively.

The data show that in the experimental scenario, the continuum robot with a fixed base is able to use the TL-FABRIKc algorithm to realize real-time solution in the process of tracking a given curve and adjusting the target end pose at the next moment. The average tracking error in the distance dimension is 16.7mm, and the average endpoint error is about 4mm when considering the 12.7mm error of the robot itself. In the pose dimension, the actual pose deviation is similar to that in \textit{Experiment} 1, which indicates that the real endpoint pose of the prototype is almost coincident with the target pose. The average distance accuracy in the virtual prototype case is 2.1mm, and the pose deviation in all directions is no more than 1°, which further indicates the reliability of the experimental results.

\begin{equation}
\label{1-14}
\left\{ \begin{array}{c}
\begin{aligned}
& x(t)=0.2\sin (t) \\ 
 & y(t)=0.1\sin (2t) \\ 
 & {{\theta }_{rot}}=\pi /6 \\ 
\end{aligned}
\end{array} \right.
\end{equation}

\begin{figure}[!t]
\centering
\includegraphics[width=3.5in]{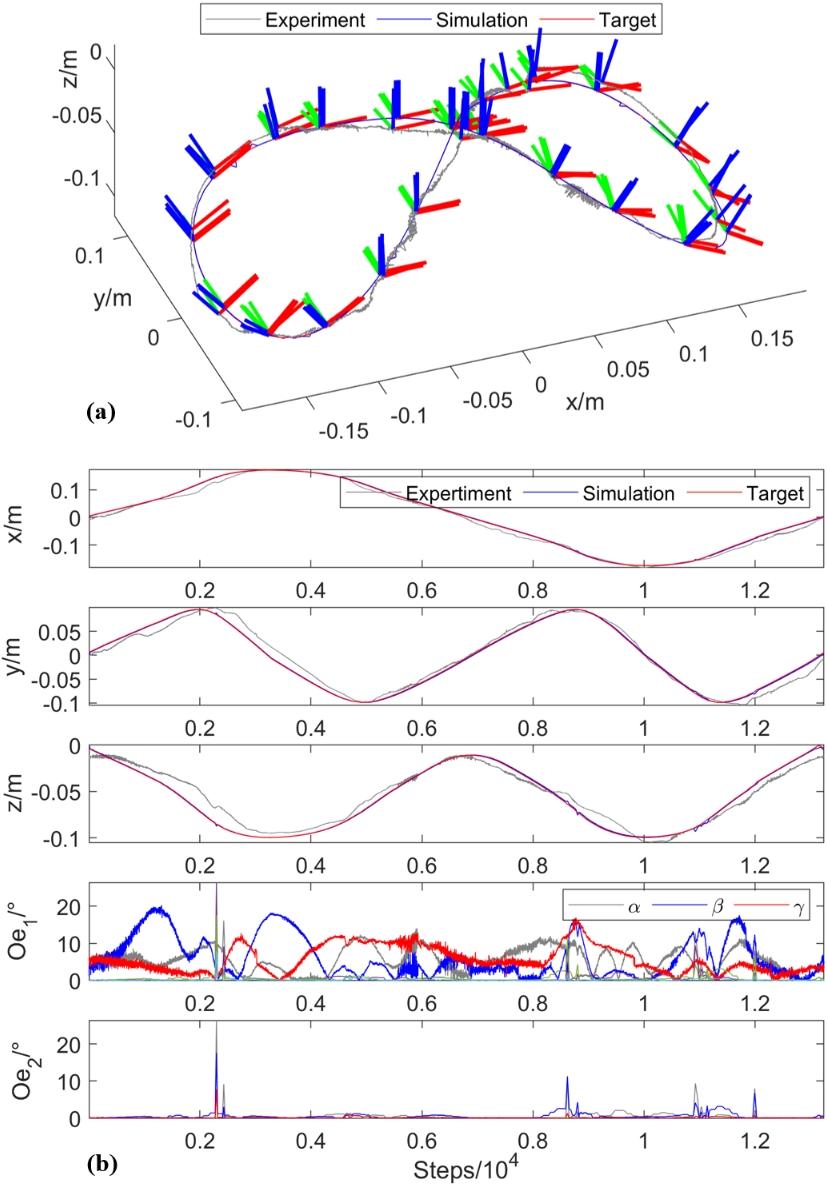}
\caption{Tip pose data analysis in \textit{Experiment} 2. (a) Tip trajectory. (b) Pose error.}
\label{figure16}
\end{figure}

\subsection{Experiment 3: The Follow-the-leader Experiment for Floating-base Continuum Robot}
\label{Section5.3}

\begin{figure}[!b]
\centering
\includegraphics[width=3.1in]{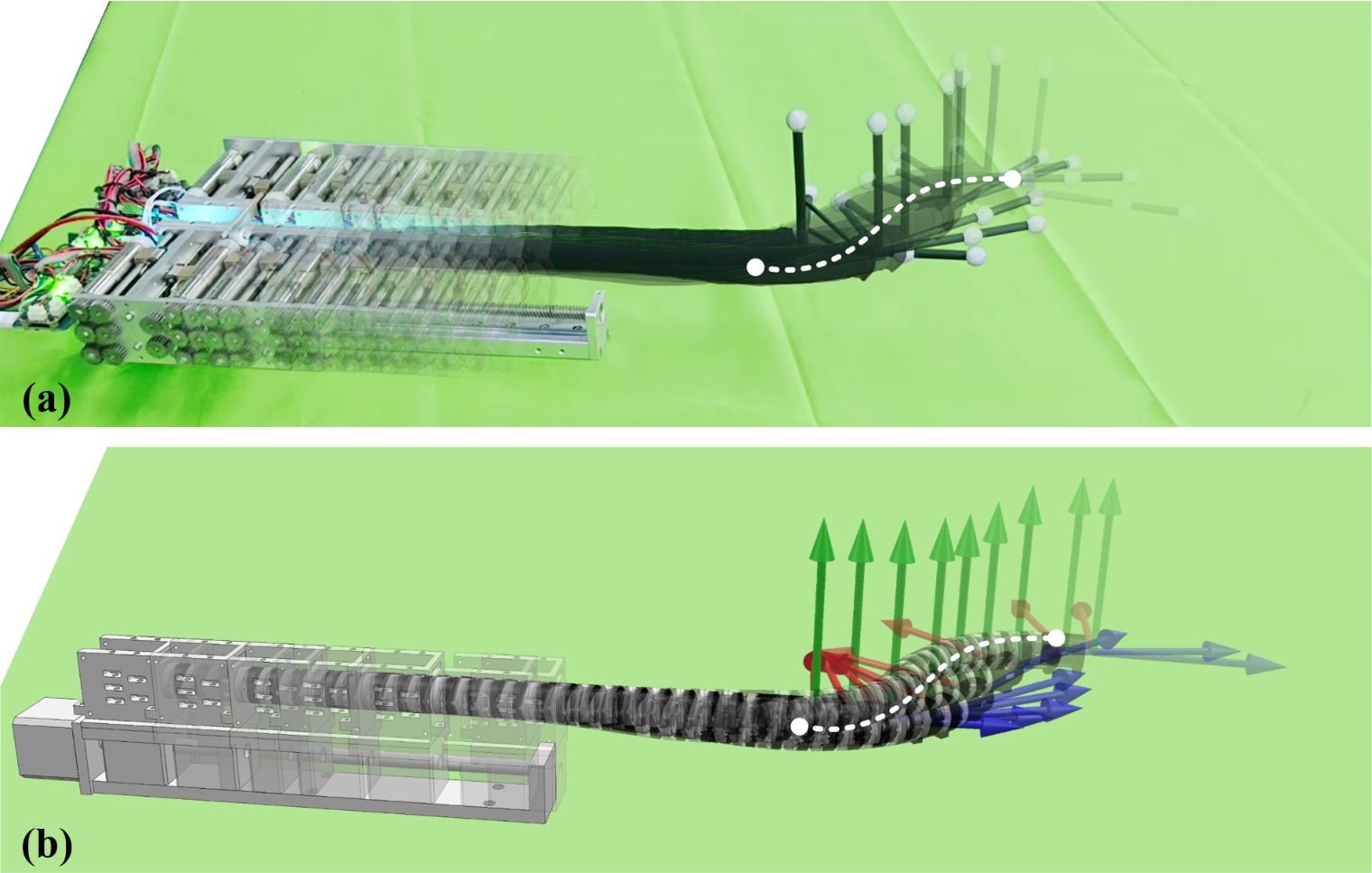}
\caption{Motion sequence in \textit{Experiment} 3. (a) FTL experiment. (b) FTL simulation in CoppeliaSim.}
\label{figure17}
\end{figure}

The purpose of \textit{Experiment} 3 is to verify the accuracy and effectiveness of the TL-FABRIKc-FTL algorithm in following a predetermined trajectory with a floating robot base. For this purpose, a planar “S” trajectory is designed along the robot end in the experiment, and the parameters of this trajectory are shown in Equation(19), the trajectory is in the same plane, which is formed by two quarter arcs tangentially connected. In Fig.\ref{figure17}, both the physical and virtual prototypes of the robot successfully realize the tracking motion of the predetermined trajectory, and the overall arm shape has a high degree of coincidence with the predetermined end trajectory. 

\begin{figure}[!t]
\centering
\includegraphics[width=3.1in]{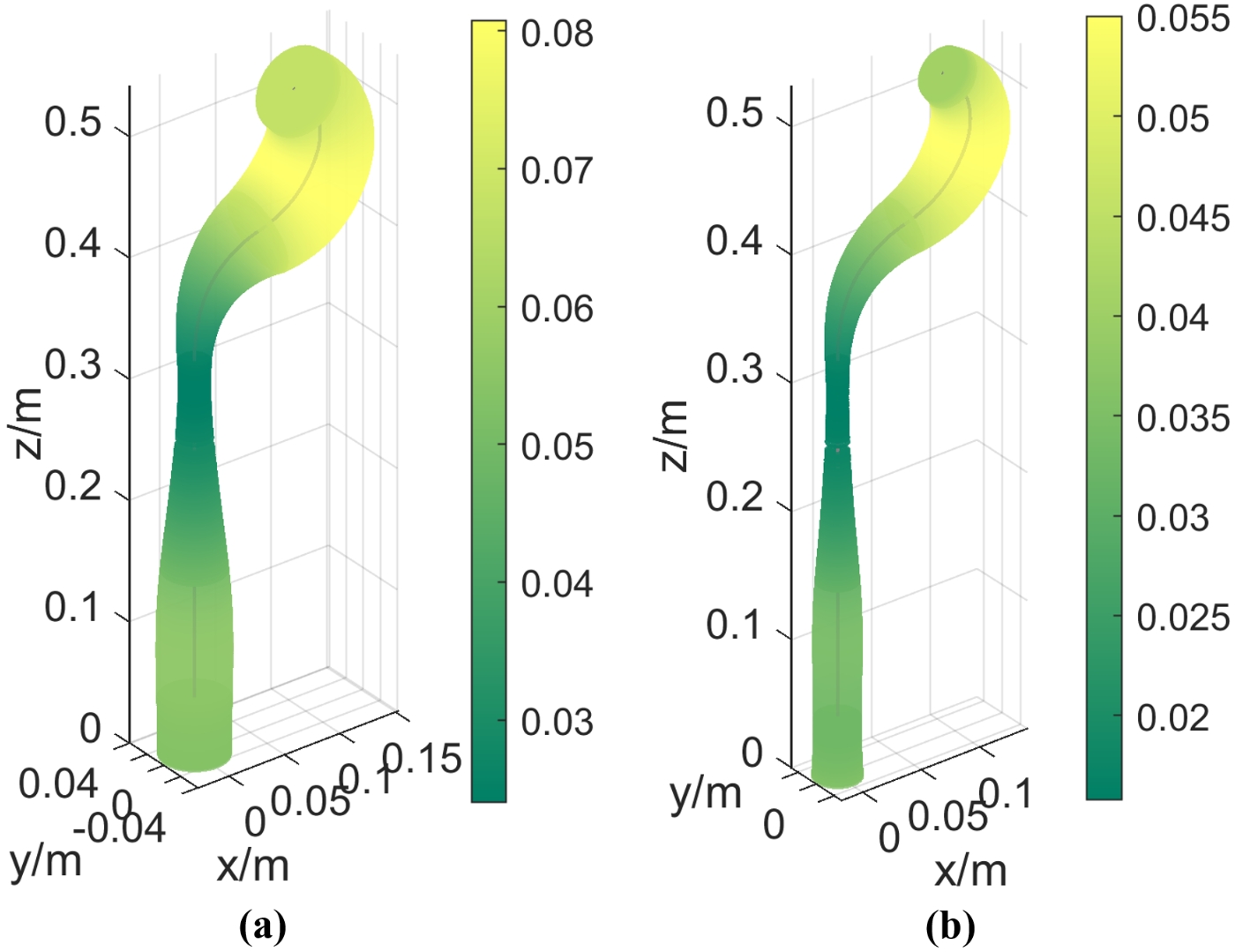}
\caption{Arm shape overlap data analysis in \textit{Experiment} 3. (a) FTL experiment. (b) FTL simulation in CoppeliaSim.}
\label{figure18}
\end{figure}

The degree of overlap between the arm shape and the trajectory during robot motion is depicted in Fig.\ref{figure18}, and with the same measurements in Section \ref{Section4.3}, the average tracking error of the physical prototype is 51.51 mm in terms of the degree of arm shape deviation, and this value is 33.15 mm in terms of the virtual prototype, and considering the robot prototype own accuracy, the average tracking error of the physical prototype shape is 38.81 mm. Though, it can be seen from Fig.\ref{figure17} that the TL-FABRIKc-FTL algorithm achieves the expected results in realizing the follow-the-leader motion of a continuum robot with a moving base, the average tracking error reflected in Fig.\ref{figure18} is large compared with the results of the numerical simulation in Section \ref{Section4.3}. Combined with the raw accuracy data of the prototype given in \textit{Experiment} 1, the true tracking accuracy of the arm shape is also still on the high side. Comparing the experimental and numerical simulation scenarios, it is analyzed that the main reason for this large error comes from the less degrees of freedom of the base. Since the mobile base in the physical prototype has only one degree of freedom of expansion and contraction, there is a situation in which the preset trajectory requirements cannot be met during the adjustment process of the arm shape, which leads to a large error in the tracking of the arm shape.

\begin{equation}
\label{1-14}
\left\{ \begin{array}{c}
\begin{aligned}
& {\theta}_{init} =[0,0] \\ 
 & {\varphi}_{init} =[0,0] \\ 
 & {{r}_{tra}}=106mm,{{l}_{tra}}=250mm \\  
\end{aligned}
\end{array} \right.
\end{equation}

\section{Conclusion}
\label{Chapter6}

This paper explores a novel TL-FABRIKc algorithm, based on a geometric iterative strategy, for solving inverse kinematics and motion planning problems in continuum robots with floating bases. First, a wire-driven continuum robot with a base that has telescopic degrees of freedom is introduced, and its effective workspace is analyzed from the perspective of its motion mechanism. Then, addressing problems in the current geometric iterative strategies used in continuum robots, such as dependence on initial arm configuration, susceptibility to local oscillations, and limitations due to fixed bases, we propose an improved two-layer geometric iteration-based TL-FABRIKc algorithm. This algorithm improves the time efficiency and accuracy in solving inverse kinematics for multi-segment continuum robots compared to similar methods. The method is further extended to scenarios involving follow-the-leader tasks under environment constraint, leading to the development of the TL-FABRIKc-FTL algorithm to accommodate floating base requirements, thereby broadening the adaptability of the continuum robot. The proposed algorithms are first validated through numerical case studies in a simulated environment. The results demonstrate that the TL-FABRIKc algorithm achieves an average accuracy improvement of 8.9\%, with 127.4 fewer iterations and a reduction of 72.6 ms in computation time compared to similar methods for solving inverse kinematics in multi-segment continuum robots. The TL-FABRIKc-FTL algorithm achieves follow-the-leader tasks with an average tracking error of within 4.06 mm. In addition, experimental scenarios were designed to verify the application of the proposed algorithms on a prototype robot. The experimental results show that TL-FABRIKc and its extended algorithm can realize the inverse kinematics solution for multi-segment continuum robots in different scenarios respectively, in which the average error at the end is about 4mm and the pose deviation in each direction is no more than 1°, which is consistent with the numerical simulation results. However, due to the limited degrees of freedom of the floating base of the physical prototype, the overall arm tracking error is large, with an average error of 38.81 mm. In summary, the TL-FABRIKc and its extended algorithm proposed in this paper is an effective method for solving the motion planning problem of multi-segmented continuum robots with floating bases, which has a better solution efficiency and effect than similar methods, and it is able to be deployed and applied in the robotic prototype. The method can be deployed and applied on robot prototypes. Meanwhile, experiments show that the degrees of freedom of the floating base have a direct effect on the implementation of the algorithm, and more degrees of freedom of the floating base will result in better arm shape tracking accuracy.

\section*{Declarations}
\begin{itemize}
\item Funding:
Supported by the  National Natural Science Foundation of China under Grant (62403340), Postdoctoral Fellowship Program of CPSF (GZC20231783) and Scientific and Technical Programs of Sichuan Province of China under Grants 2023NSFSC0475.
\item Conflict of interest:
The authors have no relevant financial or non-financial interests to disclose.
\item Data availability:
The data that support the findings of this study are available from the corresponding author upon reasonable request.
\end{itemize}


\bibliography{sn-bibliography}

\end{document}